\documentclass[acmsmall,screen]{acmart}
\usepackage{xcolor}

\usepackage{algorithm}
\usepackage{algorithmic}
\usepackage{setspace}
\usepackage{adjustbox}   
\usepackage{array,booktabs,multirow,threeparttable,adjustbox}

\usepackage{amsmath}

\let\digamma\relax

\usepackage{amssymb}
\usepackage{amsfonts}
\usepackage{algorithmic}
\usepackage{graphicx}
\usepackage{epstopdf}
\usepackage{textcomp}

\usepackage{booktabs}  
\usepackage{threeparttable}  
\usepackage{multirow}
\usepackage{footnote}
\usepackage{hyperref}
\usepackage{colortbl}
\usepackage{forest}     
\usepackage{graphicx}   

\usepackage{subfig} 

\usepackage{array}
\AtBeginDocument{%
  }

\setcopyright{acmlicensed}
\copyrightyear{2025}
\acmYear{2025}
\acmDOI{XXXXXXX.XXXXXXX}

\acmJournal{JACM}
\acmVolume{37}
\acmNumber{4}
\acmArticle{111}
\acmMonth{8}

\begin{document}

\title{Emotion Recognition from Skeleton Data: A Comprehensive Survey}

    %

\author{Haifeng Lu}
\email{luhf18@lzu.edu.cn}
\affiliation{%
  \institution{Shenzhen MSU-BIT University}
  \city{Shenzhen}
  \state{Guangdong}
  \country{China}
}

\author{Jiuyi Chen}
\email{ftchenjiuyi@mail.scut.edu.cn}
\affiliation{%
  \institution{Shenzhen MSU-BIT University}
 \city{Shenzhen}
  \state{Guangdong}
  \country{China}
}

\author{Zhen Zhang}
\email{zhangzhen19@lzu.edu.cn}
\affiliation{%
  \institution{Shenzhen MSU-BIT University}
 \city{Shenzhen}
  \state{Guangdong}
  \country{China}
}

\author{Ruida Liu}
\email{soduku645@gmail.com}
\affiliation{%
  \institution{Shenzhen MSU-BIT University}
 \city{Shenzhen}
  \state{Guangdong}
  \country{China}
}

\author{Runhao Zeng}
\authornotemark[1]
\email{zengrh@smbu.edu.cn}
\affiliation{%
  \institution{Shenzhen MSU-BIT University}
  \city{Shenzhen}
  \state{Guangdong}
  \country{China}
}

\author{Xiping Hu}
\authornote{Runhao Zeng and Xiping Hu are corresponding authors.}
	\email{huxp@bit.edu.cn}
\affiliation{%
  \institution{Shenzhen MSU-BIT University}
 \city{Shenzhen}
  \state{Guangdong}
  \country{China}
}

\renewcommand{\shortauthors}{Lu et al.}

\begin{abstract}
		Emotion recognition through body movements has emerged as a compelling and privacy-preserving alternative to traditional methods that rely on facial expressions or physiological signals. Recent advancements in 3D skeleton acquisition technologies and pose estimation algorithms have significantly enhanced the feasibility of emotion recognition based on full-body motion. This survey provides a comprehensive and systematic review of skeleton-based emotion recognition techniques. First, we introduce psychological models of emotion and examine the relationship between bodily movements and emotional expression. Next, we summarize publicly available datasets, highlighting the differences in data acquisition methods and emotion labeling strategies. We then categorize existing methods into posture-based and gait-based approaches, analyzing them from both data-driven and technical perspectives. In particular, we propose a unified taxonomy that encompasses four primary technical paradigms: Traditional approaches, Feat2Net, FeatFusionNet, and End2EndNet. Representative works within each category are reviewed and compared, with benchmarking results across commonly used datasets. Finally, we explore the extended applications of emotion recognition in mental health assessment, such as detecting depression and autism, and discuss the open challenges and future research directions in this rapidly evolving field.
\end{abstract}



\received{20 February 2007}
\received[revised]{12 March 2009}
\received[accepted]{5 June 2009}

\maketitle

\section{Introduction}\label{in}
Emotion plays a vital role in daily communication and behavioral decision-making, shaping individual cognition, social interactions, and group dynamics. In the field of Artificial Intelligence (AI), enabling machines to recognize human emotions has become a central research objective, as it significantly enhances the interactivity, adaptability, and decision-making capabilities of intelligent systems. Furthermore, emotion recognition holds substantial value across a broad spectrum of applications, including Human–Computer Interaction (HCI), mental health monitoring, educational technology, safe driving, abnormal behavior detection, intelligent tutoring systems, market analysis, and security surveillance \cite{kolakowska2014emotion, picard2000affective, garcia2017emotion}. As such, the development of efficient and accurate emotion recognition methods is critical not only for advancing AI research but also for improving the functionality and real-world applicability of intelligent systems across diverse domains.

Currently, the most widely used approaches for emotion recognition rely on facial expression analysis \cite{Li2022facialSurvey}, audio signal processing \cite{khalil2019speech}, text analysis \cite{alswaidan2020survey}, and physiological signal monitoring \cite{shu2018review, Dzedzickis2020Humansensors}. However, these methods often depend on contact-based sensors or wearable devices, which can be expensive, uncomfortable to use, and potentially intrusive, thereby raising privacy concerns and limiting their practical deployment. In summary, achieving accurate emotion recognition while preserving user privacy and ensuring a comfortable, non-invasive experience remains a significant challenge in the field.

Extensive research has shown that full-body movements play an important role in conveying emotional states, with body gestures serving as a rich, nonverbal channel for affective communication \cite{kleinsmith2012affective, noroozi2018survey, montepare1987identification, michalak2009embodiment, deligianni2019emotions}. Compared to facial expressions or voice, body-based cues—particularly those from the torso and limbs—enable long-range emotion sensing \cite{wang2023emotion}. The rapid development of depth-sensing technologies \cite{zhang2012microsoft} and advances in human pose estimation \cite{peng2024dual, zheng2020deep} have made it increasingly feasible to extract accurate 2D/3D skeleton data, which is more robust to environmental variations and better suited for privacy-sensitive scenarios than traditional visual or audio modalities. These technical advances have fueled growing interest in skeleton-based emotion recognition, an emerging topic in affective computing.

Emotion recognition based on 3D skeleton data can be broadly categorized into two main categories. The first category focuses on posture-based emotion recognition, where emotional states are inferred by analyzing motion features associated with specific actions, such as knocking, waving, or jumping. The second category centers on gait-based emotion recognition, which explores dynamic features during natural walking to model underlying emotional states. The data indicate a steady increase in the number of publications leveraging posture and gait for emotion recognition. This growth highlights the growing interest and practical potential of skeleton-based methods for emotion understanding and human–computer interaction \cite{ghaleb2021skeleton, zhang2021emotion, Beyan2023TAC}.

%
%
%
%
%
%
%

\begin{table}[tp]  
	\renewcommand\arraystretch{1.2}
	\centering  
	\fontsize{9}{12}\selectfont
	\caption{A comparison of related survey}  
	\begin{threeparttable}  
		\begin{tabular}{lccc}
			\toprule
			\textbf{} & Year & Posture & Gait \\ \midrule
			
			\rowcolor{gray!12}
			Kleinsmith et al. \cite{kleinsmith2012affective} & 2013 & $\checkmark$ & $\times$ \\
			
			Noroozi et al. \cite{noroozi2018survey} & 2018 & $\checkmark$ & $\times$ \\
			
			\rowcolor{gray!12}
			Deligianni et al. \cite{deligianni2019emotions} & 2019 & $\times$ & $\checkmark$ \\
			
			Xu et al. \cite{xu2020emotion} & 2023 & $\times$ & $\checkmark$\\
			
			\rowcolor{gray!12}
			Mahfoudi et al. \cite{9968283} & 2023 & $\checkmark$ & $\times$ \\
			
			Our & - & $\checkmark$ & $\checkmark$ \\
			
			\bottomrule
		\end{tabular}
	\end{threeparttable} 
	\label{Survey_List}
\end{table}

Despite increasing research interest in this field, a comprehensive and unified survey of skeleton-based emotion recognition is still lacking. Existing reviews tend to adopt fragmented perspectives, often overlooking a key insight: both posture and gait stem from the same underlying skeleton data and therefore share common representations, feature extraction methods, and modeling strategies. Their primary difference lies in temporal dynamics rather than data modality. As summarized in Table~\ref{Survey_List}, most existing surveys focus on a single modality—such as body expressions~\cite{kleinsmith2012affective, noroozi2018survey, 9968283} or gait-based affective analysis~\cite{deligianni2019emotions, xu2020emotion}.

Given the methodological overlap and the recent surge of interest in skeleton-based emotion recognition, it is both timely and necessary to establish a unified review framework that integrates posture- and gait-based approaches. In this work, we present a comprehensive survey that unifies posture and gait analysis under a single framework, offering a structured comparison of existing methodologies, datasets, and applications, this paper is structured in Fig. \ref{fig:survey_structure}. Sec. \ref{sec_2} introduces the fundamental emotion models and examines the relationship between body movements and emotional expression. In Sec. \ref{dataset}, we review existing public datasets, highlight variations in data collection methods, and compare the characteristics of each dataset. Sec. \ref{sec_4} provides a detailed analysis of methodologies, categorizing them into posture-based and gait-based approaches, and examines the technical features of each. Sec. \ref{sec_5} explores extended applications of skeleton-based emotion recognition. Finally, Sec. \ref{sec_6} concludes the survey and outlines promising directions for future research.

\newlength{\myMheight}        
\setlength{\myMheight}{0.4cm}   
	\begin{figure*}[tp]
		\centering
		\resizebox {0.92\textwidth}{!}{%
			\sf
			\begin{forest}
				for tree={
					grow=east,
					reversed=true,
					anchor=base west,
					parent anchor=east,
					child anchor=west,
					base=left,
					font=\normalsize,
					rectangle,
					draw=black,
					rounded corners,
					align=left,
					minimum width=4em,
					edge+={black, line width=0.5pt},
					s sep=3pt,
					l sep=8mm,
					inner xsep=4pt,
					inner ysep=1pt,
					edge path={
						\noexpand\path [draw, \forestoption{edge}]
						(!u.parent anchor) -- ++(3.5mm,0) |- (.child anchor) \forestoption{edge label};
					},
					ver/.style={rotate=90, child anchor=north, parent anchor=south, anchor=center},
				},
where level=1{ text width=13em, font=\fontsize{10pt}{12pt}\selectfont }{},
where level=2{ text width=21em, font=\fontsize{10pt}{12pt}\selectfont }{},
where level=3{ text width=18em, font=\fontsize{10pt}{12pt}\selectfont }{},
				[Emotion Recognition from Skeleton Data: A Comprehensive Survey, ver,
				color=black, fill=orange!40, text=black, text width=38em, text centered
				[{\includegraphics[width=\myMheight]{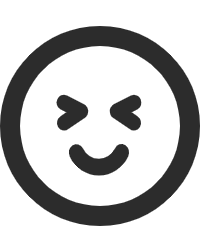}}
				Models of Emotion \\ and Body Movement (Sec.\ref{sec_2}),
				color=black, fill=blue!13, text=black
				[Models of Emotion (Sec.\ref{sec_2_1}),
				color=black, fill=blue!8, text=black
				[Discrete Emotions Theory, color=black, fill=blue!5]
				[Multidimensional Emotions Theory, color=black, fill=blue!5]
				[Componential Emotions Theory, color=black, fill=blue!5]
				]
				[Emotional Body Expressions (Sec.\ref{sec_2_2}),
				color=black, fill=blue!8, text=black
				[Posture-based Emotion Expression, color=black, fill=blue!5]
				[Gait-based Emotion Expression, color=black, fill=blue!5]
				]
				]
				[{\includegraphics[width=\myMheight]{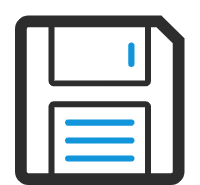}} Data (Sec.\ref{dataset}),
				color=black, fill=brown!40, text=black
				[Data Collection Paradigm (Sec.\ref{Capture_paradigm}), color=black, fill=brown!20]
				[Body Movement Capture Methodology (Sec.\ref{Capture_Methodology}),
				color=black, fill=brown!20, text=black
				[Optical Motion Capture System, color=black, fill=brown!10]
				[Inertial Motion Capture System, color=black, fill=brown!10]
				[Depth-sensing Camera, color=black, fill=brown!10]
				[RGB Camera, color=black, fill=brown!10]
				]
				[Posture-based Datasets (Sec.\ref{Posture_data}), color=black, fill=brown!20]
				[Gait-based Datasets (Sec.\ref{gait_data}), color=black, fill=brown!20]
				[Dataset Comparison and Analysis (Sec.\ref{Dataset_con}), color=black, fill=brown!20]
				]
				[{\includegraphics[width=\myMheight]{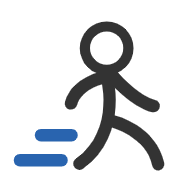}}
				Body Movement Based \\Emotion Recognition (Sec.\ref{sec_4}),
				color=black, fill=olive!25, text=black
				[Posture-based Emotion Recognition (Sec.\ref{posture_based_method}),
				color=black, fill=olive!15, text=black
				[Traditional Approaches, color=black, fill=olive!5]
				[Feat2Net Approaches, color=black, fill=olive!5]
				[End2EndNet Approaches, color=black, fill=olive!5]
				[Pre-training Approaches, color=black, fill=olive!5]
				[Evaluation of Methods on Public Datasets, color=black, fill=olive!5]
				]
				[Gait-based Emotion Recognition (Sec.\ref{gait_based_method}),
				color=black, fill=olive!15, text=black
				[Traditional Approaches, color=black, fill=olive!5]
				[Feat2Net Approaches, color=black, fill=olive!5]
				[FeatFusionNet Approaches, color=black, fill=olive!5]
				[End2EndNet Approaches, color=black, fill=olive!5]
				[Unsupervised Approaches, color=black, fill=olive!5]
				[Evaluation of Methods on Public Datasets, color=black, fill=olive!5]
				]
				[Methods Comparison and Analysis (Sec.\ref{sec4_3}),color=black, fill=olive!15, text=black]
				]
				[{\includegraphics[width=\myMheight]{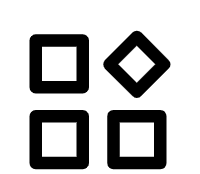}}Task-Specific Applications \\ (Sec.\ref{sec_5}),
				color=black, fill=orange!25, text=black
				[Depression Detection Using Skeleton \\Data (Sec.\ref{sec_5_1}), color=black, fill=orange!15]
				[Autism Detection Using Skeleton \\Data (Sec.\ref{sec_5_2}), color=black, fill=orange!15]
				[Abnormal Behavior Detection Using Skeleton \\Data (Sec.\ref{sec_5_3}), color=black, fill=orange!15]
				]
				[{\includegraphics[width=\myMheight]{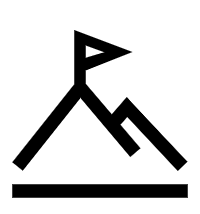}} Challenges and Future \\Research Directions (Sec.\ref{sec_6}),
				color=black, fill=pink!40, text=black,
				[Constructing Diversified Datasets (Sec.\ref{sec_6_1}), color=black, fill=pink!30,text width=33em]
				[Improving Model Performance (Sec.\ref{sec_6_3}), color=black, fill=pink!30,text width=33em]
				[Building End-to-End and Efficient Emotion Recognition Frameworks (Sec.\ref{sec_6_4}), color=black, fill=pink!30,text width=33em]
				[Expanding to Multi-person Emotion Recognition (Sec.\ref{sec_6_5}), color=black, fill=pink!30,text width=33em]
				[Expanding to Multimodal Emotion Recognition (Sec.\ref{sec_6_6}), color=black, fill=pink!30,text width=33em]
				[Leveraging Large Models for Skeleton-Based Emotion Recognition (Sec.\ref{sec_6_7}), color=black, fill=pink!30,text width=33em]
				]
				]
			\end{forest}%
		}
		\caption{The overall structure of our survey}
		\label{fig:survey_structure}
	\end{figure*}


\section{Models of Emotion and Body Movement}\label{sec_2}

This section provides a comprehensive overview of the major emotion models discussed in Sec. \ref{sec_2_1}, including discrete, dimensional, and componential models. The relationship between emotional states and posture-based expressions is reviewed in Sec. \ref{sec_2_2_1}, while gait-based emotional expressions are explored in Sec. \ref{sec_2_2_2}.

\subsection{Models of Emotion}\label{sec_2_1}

From a psychological perspective, emotions are stimulus-driven responses characterized by distinct physiological changes \cite{rached2013emotion}, and are commonly classified as reactional, hormonal, or automatic processes \cite{dalgleish2004emotional}. The modeling of affect has long been a subject of scholarly debate, with three prominent approaches emerging: discrete, dimensional, and componential models \cite{kolakowska2015modeling}.

\subsubsection{Discrete Emotions Theory}\label{sec_2_1_1}

The classification of emotions into distinct, easily recognizable categories has long been a foundational approach in emotion research. A widely accepted perspective, heavily influenced by the work of Paul Ekman \cite{ekman1971universals}, posits the existence of a set of universal primary emotions—typically happiness, sadness, fear, anger, disgust, and surprise—that are biologically hardwired and universally recognized across cultures. This discrete-state view has gained significant traction in affective computing due to its conceptual simplicity and its claim of cross-cultural applicability.

However, the discrete emotions theory has faced increasing debate. While the discrete emotions theory emphasize emotional consistency across cultures, numerous studies have demonstrated that emotional perception and expression are significantly influenced by factors such as age, gender, and cultural or linguistic context \cite{gunes2011emotion, stephens2017automatic, kleinsmith2012affective}. These factors are critical for experimental design and the accurate interpretation of findings across diverse populations, and have been extensively discussed in the literature.

\subsubsection{Multidimensional Emotions Theory}\label{sec_2_1_2}

Another widely adopted approach to emotion modeling is the dimensional model, which represents emotions as points within a continuous, multidimensional space \cite{russell1977evidence, greenwald1989affective}. Among the most commonly used dimensions are valence (the degree of pleasantness or unpleasantness) and arousal (the level of activation or intensity). This framework acknowledges the complexity of emotional experiences and enables more nuanced analysis by capturing subtle variations across emotional states.
            
\begin{figure}[tp]
    \centering
    \subfloat[The 2D VA emotion model]{\includegraphics[width=1.5in]{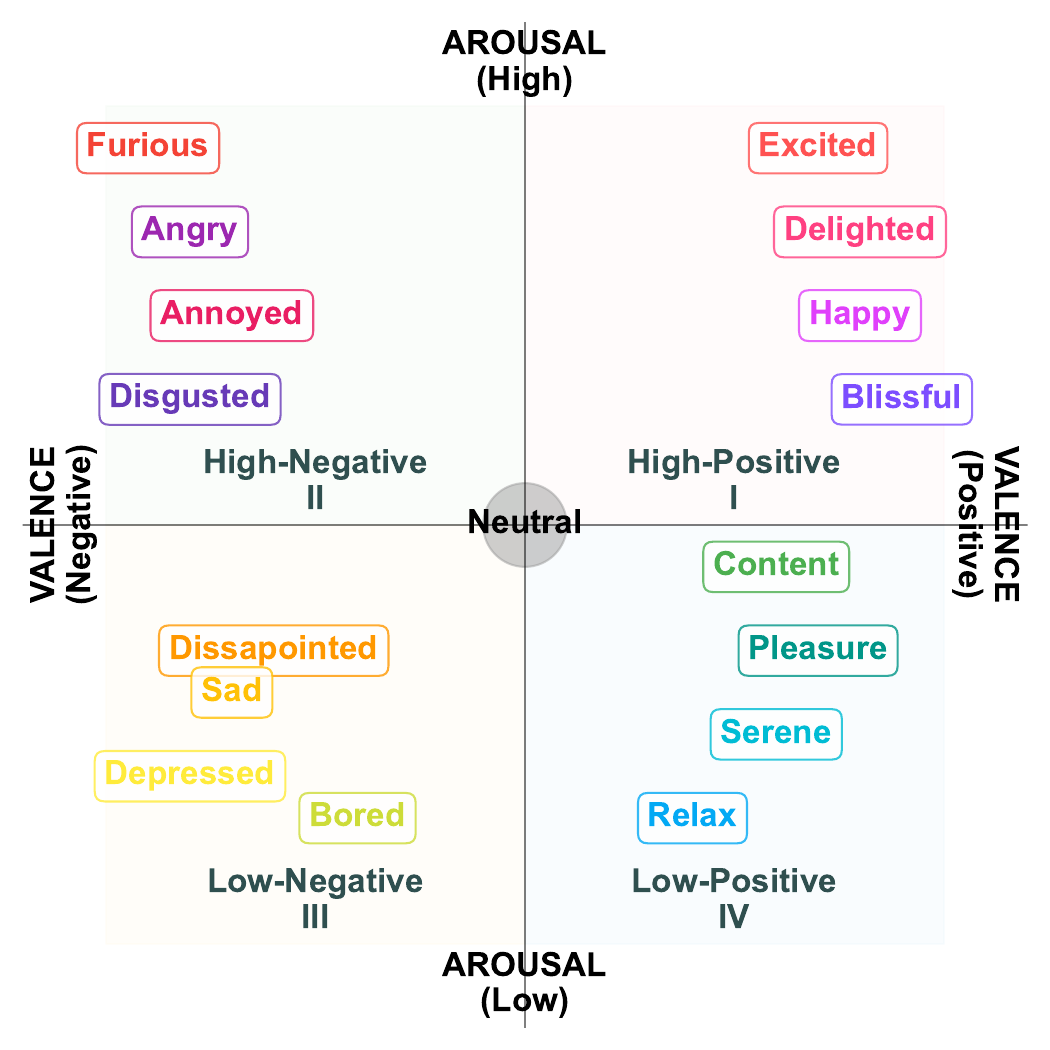}} \hspace{0.2cm}
    \subfloat[The 3D PAD emotion model]{\includegraphics[width=1.8in]{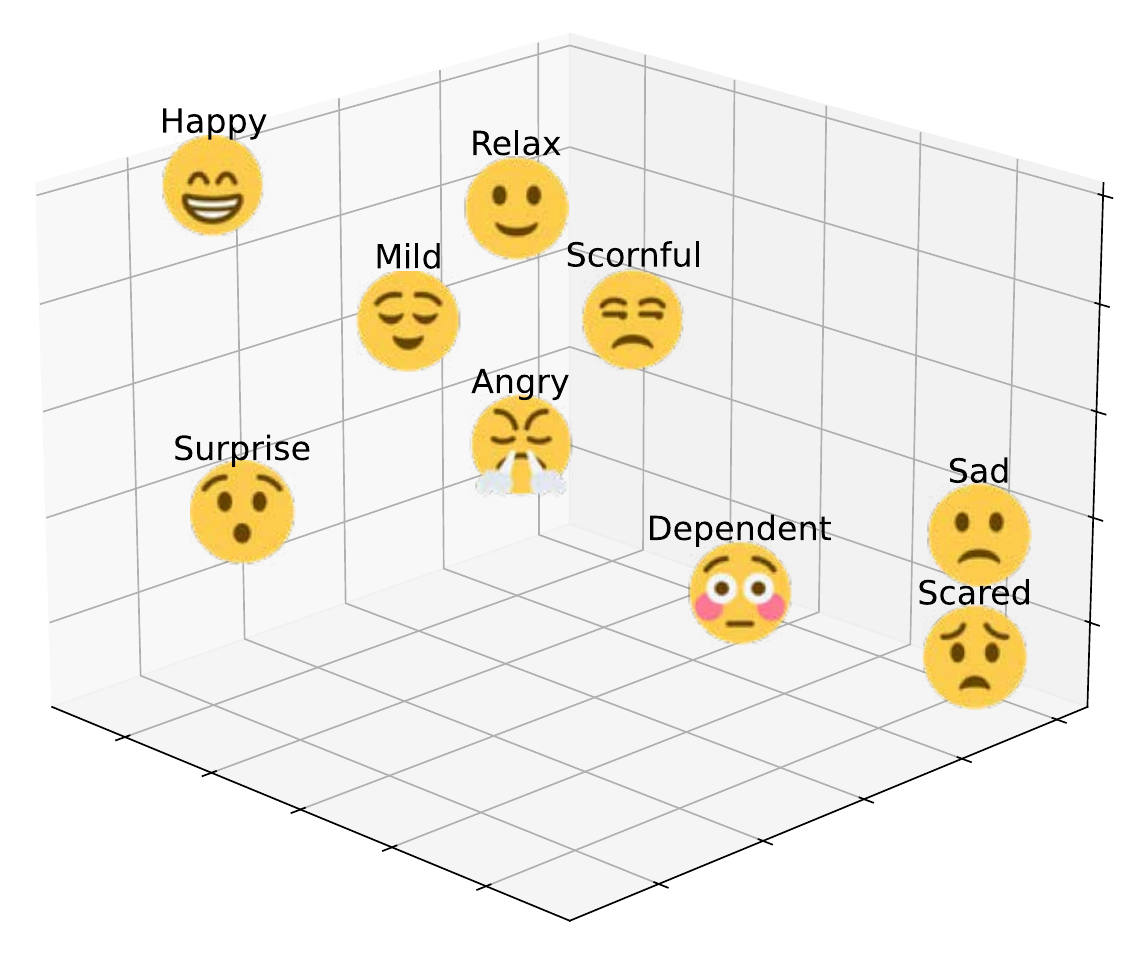}} \hspace{0.2cm}
    \subfloat[The Plutchik’s model]{\includegraphics[width=1.5in]{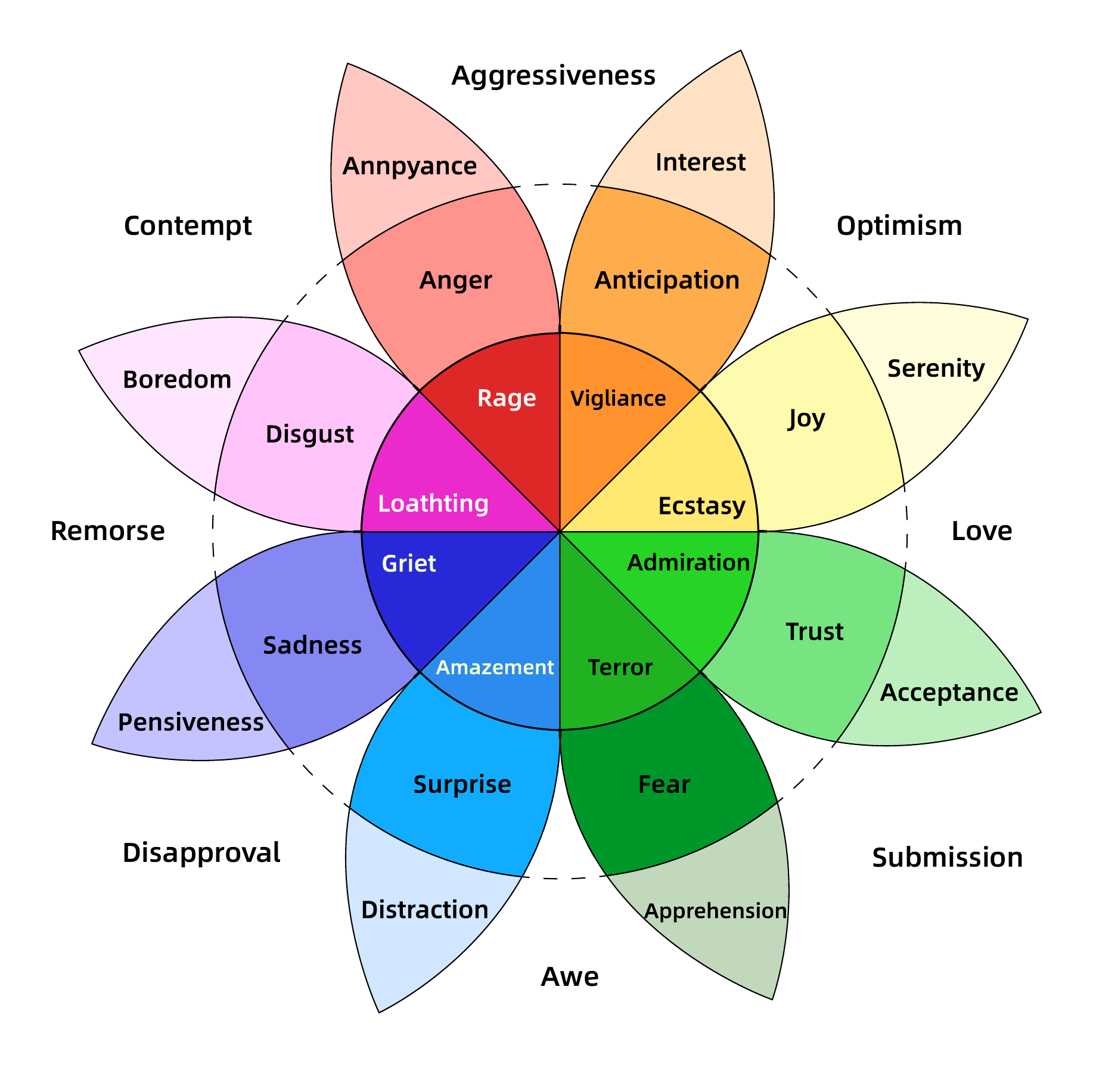}}
    \caption{Various emotion models}
    \label{emotion_models}
\end{figure}

The 2D VA model (Valence–Arousal), most notably represented by Russell’s Circumplex Model \cite{wilson2003real} (see Fig. \ref{emotion_models}.(a)), maps emotions onto the valence–arousal plane and has been widely applied across disciplines such as psychology, marketing, and education. Similarly, the 3D PAD model (Pleasure, Arousal, Dominance), proposed by Mehrabian \cite{mehrabian1996pleasure} (see Fig. \ref{emotion_models}.(b)), extends this framework by incorporating dominance as a third axis. These continuous models offer a rich descriptive space for representing emotions and have been shown to relate intuitively to physiological signals, particularly those associated with valence and arousal \cite{stephens2017automatic, kleinsmith2012affective}. Notably, studies analyzing body movements in relation to affect suggest that the arousal dimension accounts for the greatest variance in movement-related emotional expression \cite{kang2016effect, kang2015emotional, barliya2013expression}.

However, despite their theoretical richness, dimensional models present challenges for automatic emotion recognition systems, as mapping subtle and continuous affective states to discrete, observable body expressions remains a non-trivial task.

\subsubsection{Componential Emotions Theory}\label{sec_2_1_3}

Componential models occupy a middle ground between categorical and dimensional approaches in terms of descriptive capacity. These models organize emotions hierarchically, positing that complex emotions arise from combinations of more basic ones. A well-known example is Plutchik’s model (see Fig. \ref{emotion_models}.(c)) \cite{plutchik2001nature}, which defines complex emotions as dyads—pairs of basic emotions—whose likelihood decreases as their complexity increases.

Although less prevalent in affective computing, compound emotions such as "happily surprised" or "angrily surprised" have garnered growing interest for their greater expressive flexibility. Componential models offer a useful balance between interpretability and richness, making them a promising framework for developing more discriminative affective computing systems.


\subsection{Emotional Body Expressions}\label{sec_2_2}

\subsubsection{Posture-based Emotion Expression}\label{sec_2_2_1}
                                                            
Human emotions are closely intertwined with bodily movements, as emotional states often manifest through subtle yet consistent changes in physical behavior \cite{wallbott1998bodily}. Scientific studies have demonstrated that parameters such as joint speed, body sway, and gesture dynamics are modulated by affective states. These observable variations in movement serve as important non-verbal cues that can be analyzed to infer an individual’s emotional condition.

In the investigation of posture and emotional expression, Dahl et al. \cite{dahl2007visual} proposed that musicians convey core emotions through distinct movement patterns: happiness and anger are expressed via large, fast gestures (fluent versus jerky), sadness is characterized by small, slow, smooth motions, and fear is marked by minimal, jerky movements with low recognizability due to potential suppression. In \cite{gross2010methodology}, the author observed that emotion-specific movement patterns vary across affective states: anger involves large, forceful gestures and rapid striking; sadness is expressed through slower, constricted motions with reduced range; joy features fluent, frequent movements and an upward head tilt; and anxiety is characterized by tense, restricted actions and a hurried rhythm. Dael et al. \cite{dael2012emotion} further identified that emotion-specific movement patterns are manifested through distinct bodily expressions: anger is associated with a forward-leaning posture, outstretched arms, and aggressive gestures; amusement includes intermittent behaviors such as object-touching and an upright, sideways-oriented head directed toward the interlocutor; and joy often involves an upward and off-centered head tilt accompanied by asymmetric arm movements.

To better illustrate the strong association between body movement patterns and discrete emotional states, this paper summarizes the characteristic motion features corresponding to four basic emotions, as presented in Table \ref{tab:Emotion_Posture_feature}.

\newcolumntype{L}[1]{>{\raggedright\arraybackslash}m{#1}}
\newcolumntype{C}[1]{>{\centering\arraybackslash}m{#1}}
\newcolumntype{P}[1]{>{\arraybackslash}m{#1}}

\begin{table}[tp]
	\centering
	\fontsize{9}{12}\selectfont
	\renewcommand\arraystretch{1.15}
	\setlength{\tabcolsep}{0pt}
	\caption{Typical posture characteristics under different emotional states}
	
	\begin{adjustbox}{max width=\textwidth}
		\begin{tabular}{
				L{2.3cm}
				P{9.5cm}
				C{4.0cm}
			}
			\toprule
			\multicolumn{1}{c}{Emotion} &
			\multicolumn{1}{c}{Posture Characteristics} &
			\multicolumn{1}{c}{3D Display} \\
			\midrule
			\rowcolor{gray!12}
			Happiness &
			Open and dynamic posture with moving arms and legs, feet oriented toward objects or people of interest. Often accompanied by spontaneous, non-goal-directed actions such as jumping, clapping, stamping, or rhythmic body movements. The body is typically held upright and may shake during intense laughter. &
			\includegraphics[width=3.5cm,trim=2pt 2pt 2pt 2pt,clip]{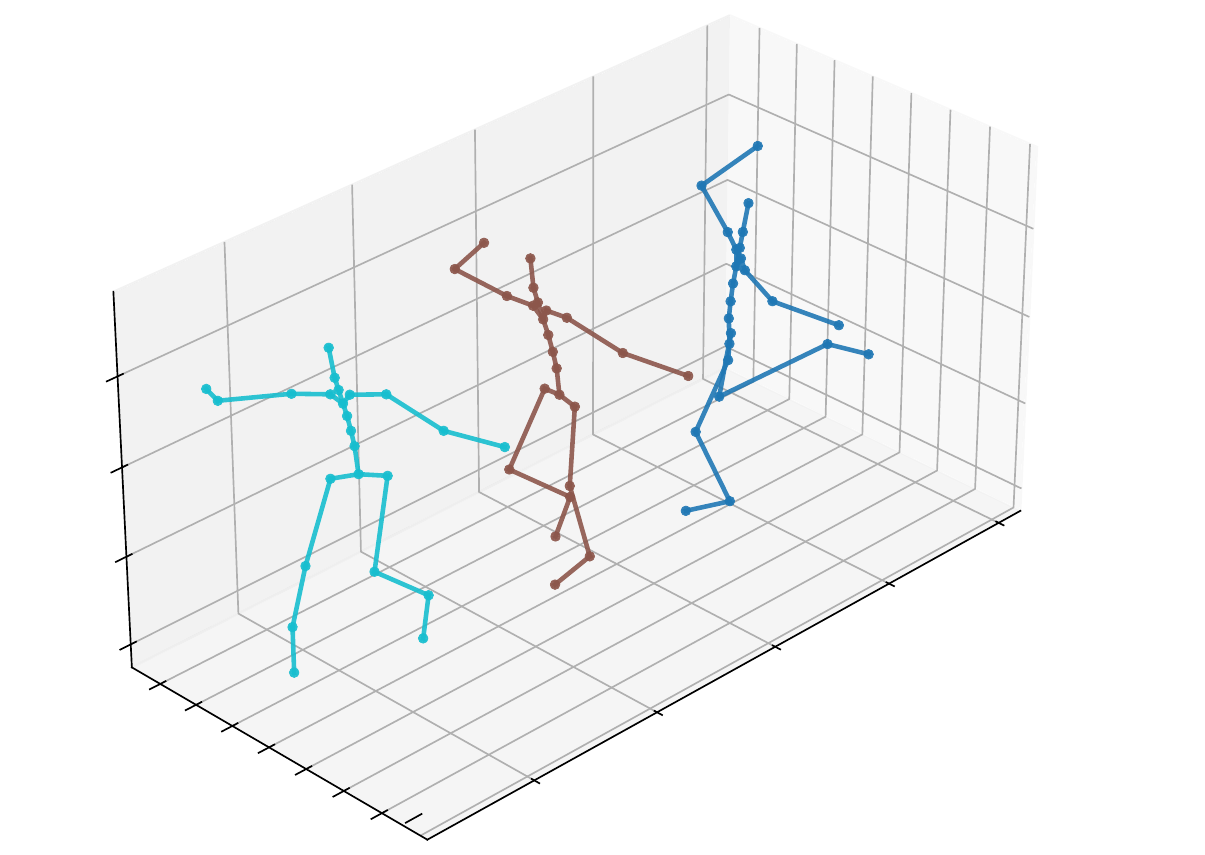} \\
			
			Anger &
			Expansive posture with hands on hips or waist, clenched fists, palm-down or pointing gestures. Frequently includes trembling or shaking of hands or body, squared shoulders, erect head, and expanded chest. The stance is firm, with rigid or suspended arms. Aggressive actions such as pacing, striking objects, or making frantic, purposeless gestures may occur. &
			\includegraphics[width=3.5cm,trim=2pt 2pt 2pt 2pt,clip]{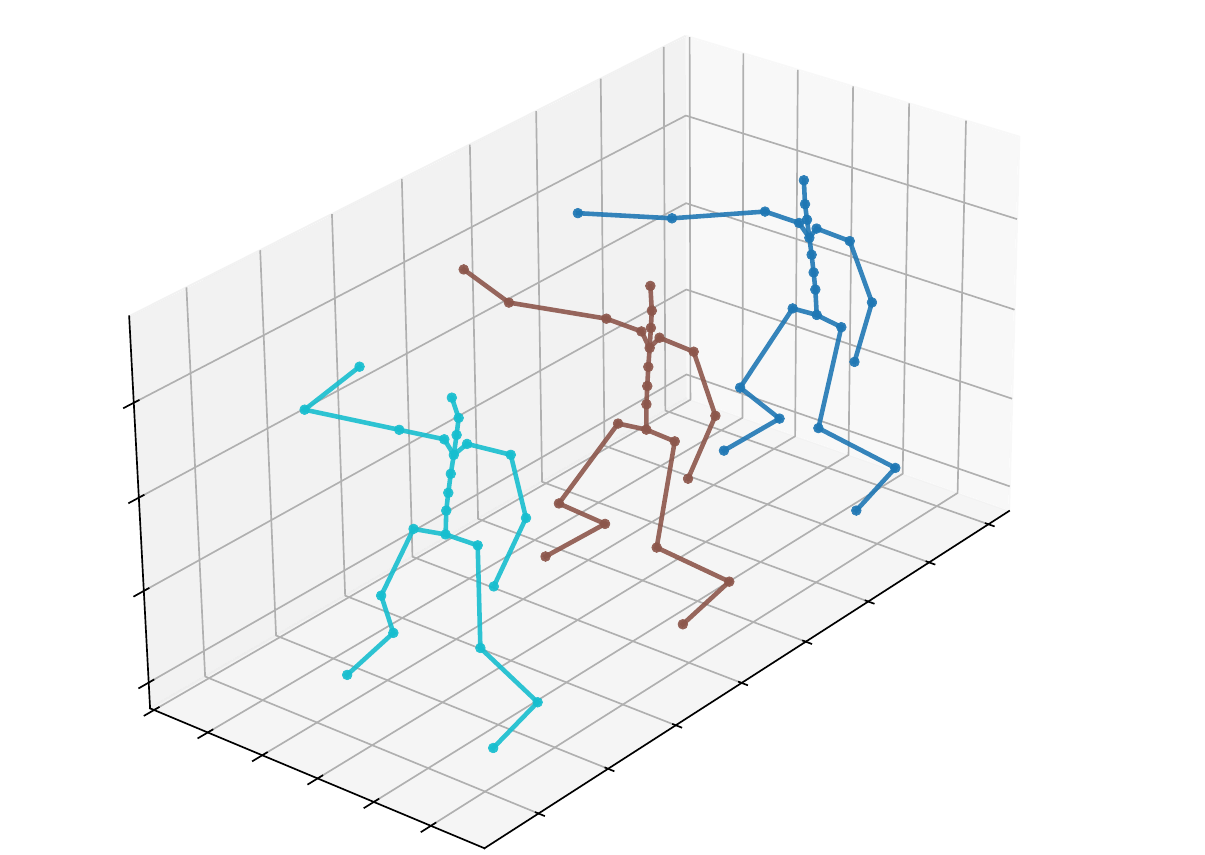} \\
			
			\rowcolor{gray!12}
			Sadness &
			Contracted posture with bowed shoulders, forward-leaning trunk, and a lowered or motionless head. Hand movements are slow or minimal, often involving self-touching of the face, head, or neck. Overall body language is passive and withdrawn, indicating emotional withdrawal. &
			\includegraphics[width=3.5cm,trim=2pt 2pt 2pt 2pt,clip]{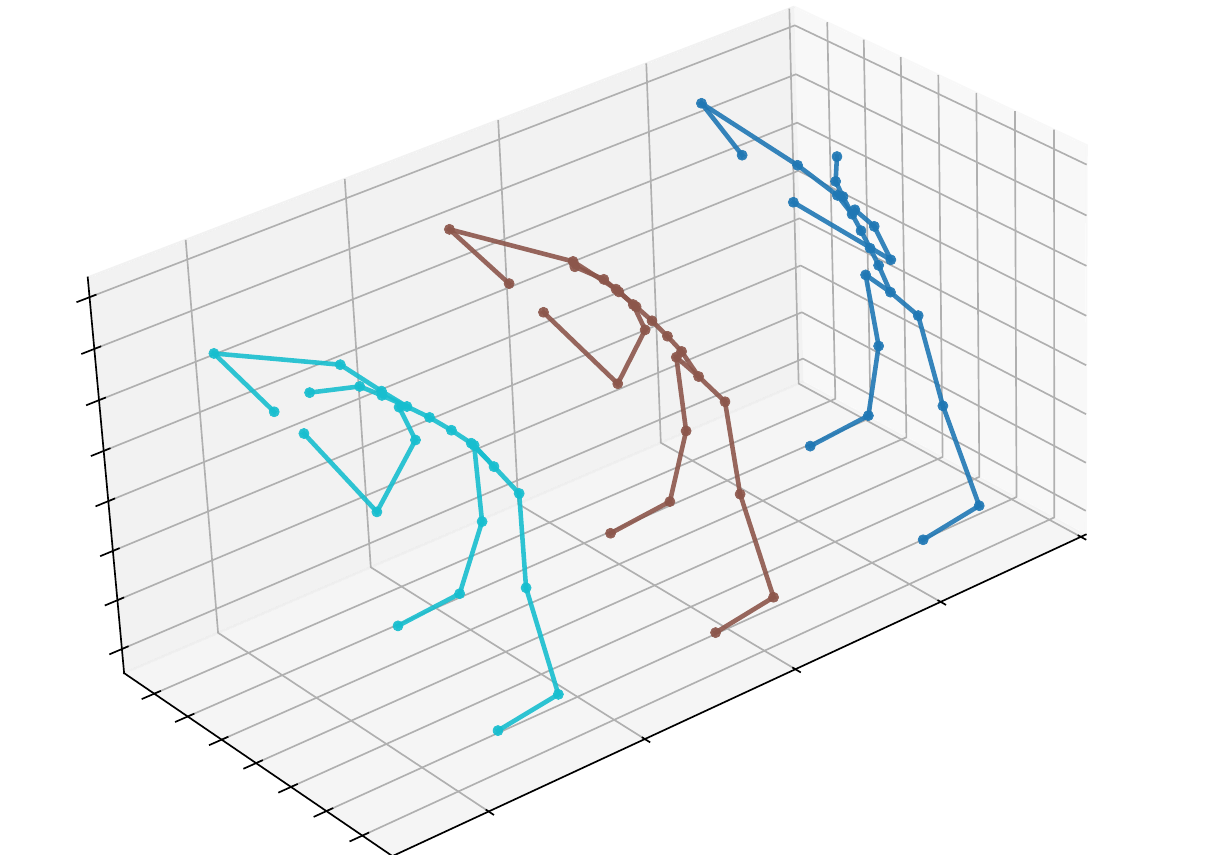} \\
			
			Fear &
			Postures characterized by crossed limbs and visible muscular tension, such as clenched hands and inward-drawn elbows. Movements may be restrained, sudden, or bouncy. The head is typically lowered between the shoulders, or the body may adopt a crouched, motionless stance. &
			\includegraphics[width=3.5cm,trim=2pt 2pt 2pt 2pt,clip]{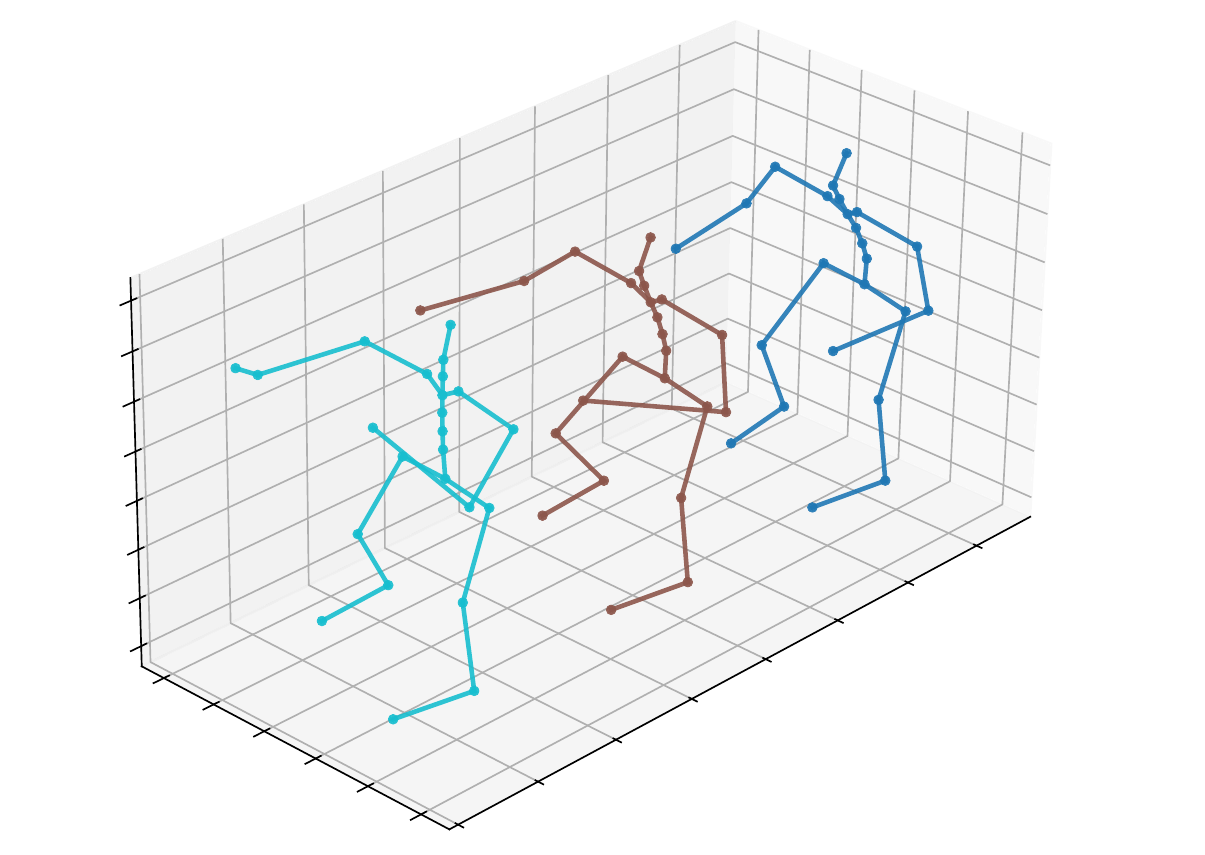} \\
			\bottomrule
		\end{tabular}
	\end{adjustbox}
	
	\label{tab:Emotion_Posture_feature}
\end{table}

\subsubsection{Gait-based Emotion Expression}\label{sec_2_2_2}

Gait refers to the periodic locomotion pattern exhibited by humans during walking and serves as an important biometric trait. Studies have shown that emotional states can significantly influence the kinematic parameters of gait, including walking speed, acceleration, and other movement features associated with specific emotions \cite{deligianni2019emotions}.

In the investigation of gait and emotional expression, Michalak et al. \cite{michalak2009embodiment} found that individuals tend to display a more bouncy gait with increased arm swing when experiencing happiness. In contrast, sadness results in slower, heavier steps, reduced arm movement, and a more relaxed upper body posture. Anger is typically characterized by fast and forceful stomping, while fear is associated with a rapid but short-stepped gait. Kim et al. \cite{kim2004emotion} reported that peak plantar pressure shifts toward the forefoot under happy emotions, whereas it concentrates on the outer rear heel during sadness. Cross et al. \cite{gross2012effort} conducted kinematic analyses on motion capture data from 16 individuals walking under five emotional conditions (happiness, satisfaction, fear, anger, and neutrality). Their results showed that gait speed was highest in happy and angry states and lowest during sadness. Additionally, sad participants frequently bent their necks and contracted their chests, whereas happy individuals tended to extend their torsos or lower their shoulders. Montepare et al. \cite{montepare1987identification} examined gait patterns in ten female undergraduates across four emotional conditions. They observed that arm swing was significantly reduced during sadness, while happy emotions resulted in faster gait rhythms and more dynamic arm movements. Emotions such as anger and pride were associated with increased stride length. Furthermore, observers were able to accurately infer participants’ emotional states based on their gait features.

To more intuitively illustrate the strong correlation between gait patterns and emotional states, this paper summarizes the typical gait characteristics associated with four basic emotions, as shown in Table \ref{tab:Emotion_Gait_feature}.

\begin{table}[tp] 
	\centering
	\fontsize{9}{12}\selectfont
	\renewcommand\arraystretch{1.15}
	\setlength{\tabcolsep}{0pt}
	\caption{Typical gait characteristics under different emotional states}
	
	\begin{adjustbox}{max width=\textwidth}
		
		\begin{tabular}{
				L{2.3cm}   
				P{9.4cm}   
				C{3.3cm}   
			}
			\toprule
			\multicolumn{1}{c}{Emotion} &
			\multicolumn{1}{c}{Gait Characteristics} &
			\multicolumn{1}{c}{3D Display} \\
			\midrule
			
			\rowcolor{gray!12}
			Happiness &
			Increased gait speed and cadence, greater step length, enhanced arm swing, noticeable vertical head movement, elevated thigh angles, and overall smoother motion patterns. The torso often appears more expansive or relaxed, frequently accompanied by a bouncy walking rhythm. &
			\includegraphics[width=3.3cm,trim=2pt 2pt 2pt 2pt,clip]{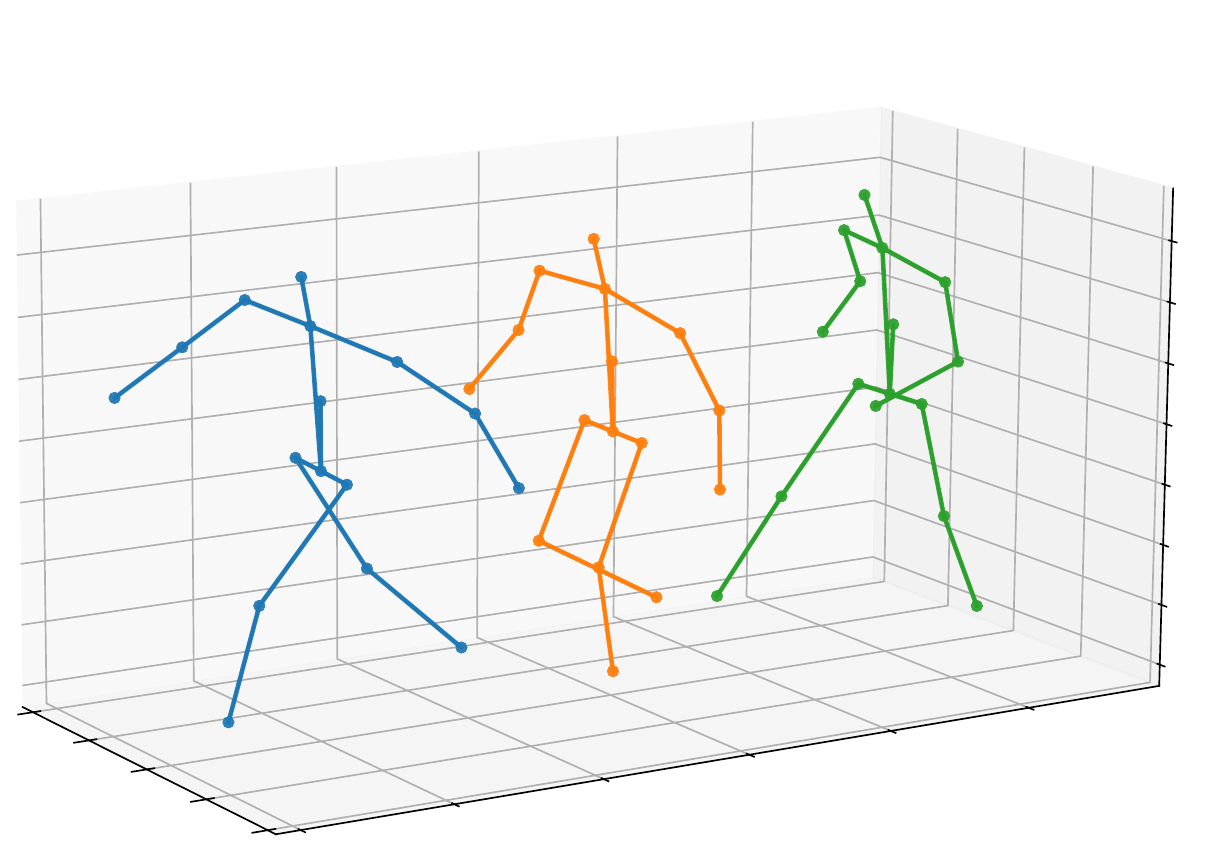} \\
			
			Anger &
			Faster gait with higher cadence, longer stride length, pronounced thigh elevation, and vigorous arm swing. Inter-segment coordination is smoother but footfalls are heavier, conveying forceful intent. &
			\includegraphics[width=3.3cm,trim=2pt 2pt 2pt 2pt,clip]{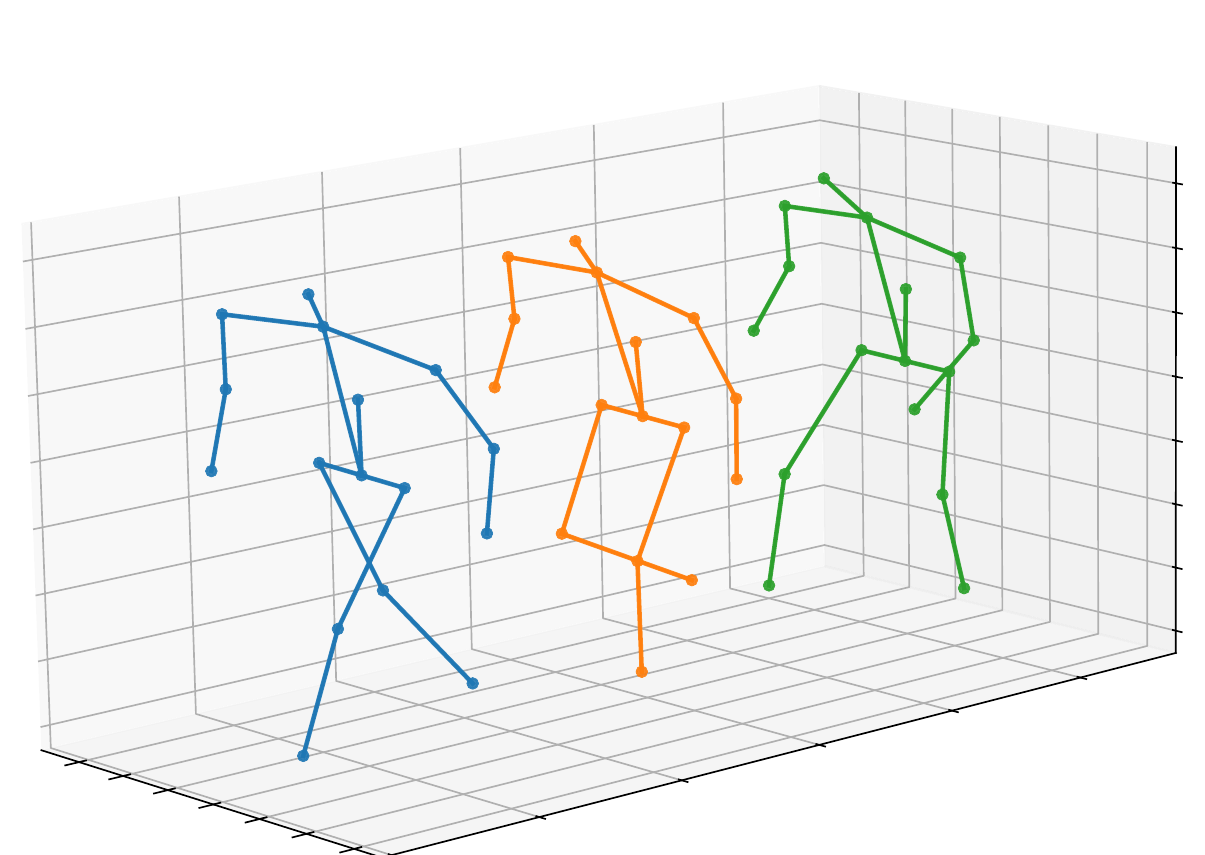} \\
			
			\rowcolor{gray!12}
			Sadness &
			Reduced walking speed and cadence, shorter strides, diminished arm swing, and limited upper-body involvement. Joint excursions—pelvic rotation, hip and shoulder flexion—are minimal. Posture tends to be slouched, with head flexion and chest contraction. &
			\includegraphics[width=3.3cm,trim=2pt 2pt 2pt 2pt,clip]{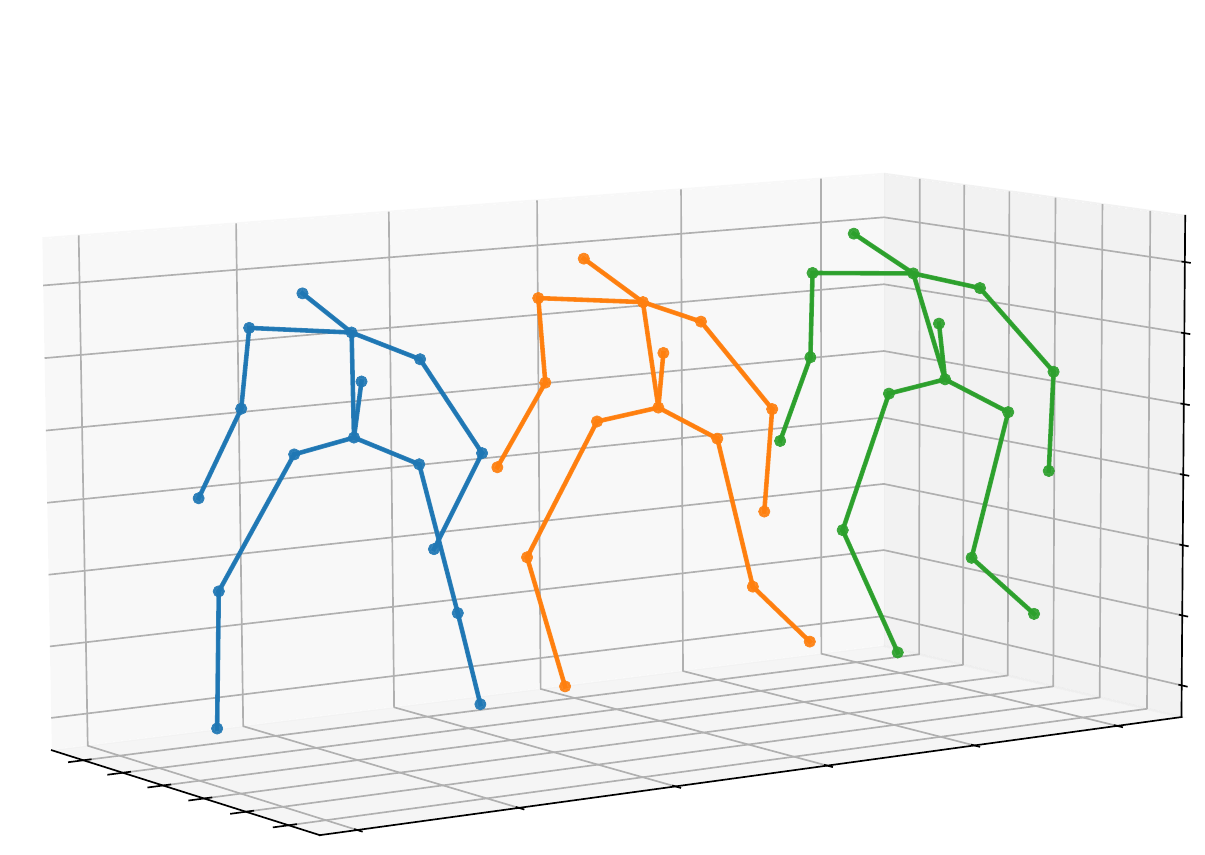} \\
			
			Fear &
			Elevated cadence with shorter step length and reduced overall speed. Posture is tense and contracted, limbs show greater flexion, and motions can be sharp or jittery. Walk often appears cautious, with increased thigh lift and guarded foot placement. &
			\includegraphics[width=3.3cm,trim=2pt 2pt 2pt 2pt,clip]{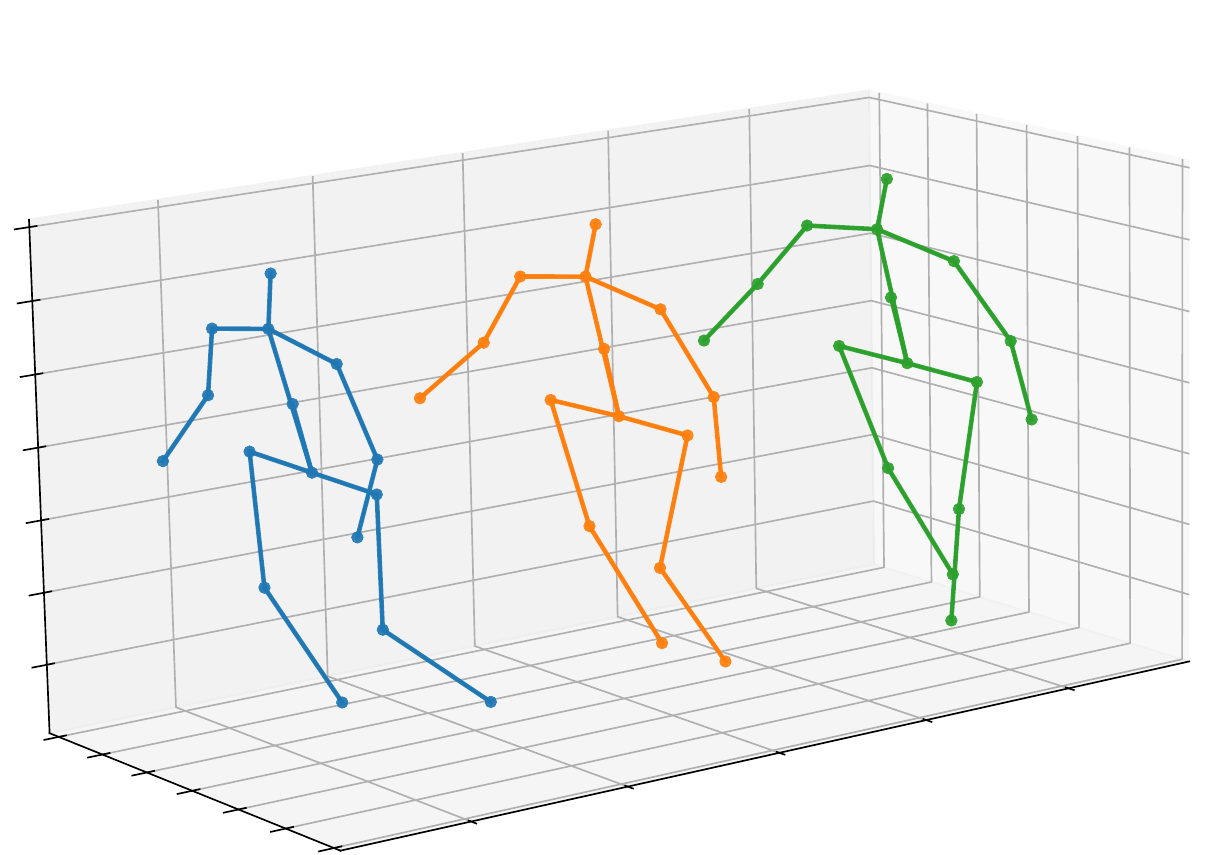} \\
			\bottomrule
		\end{tabular}
	\end{adjustbox}
	
	\label{tab:Emotion_Gait_feature}
\end{table}

\section{Data}\label{dataset}

This section provides a comprehensive overview of common methods for capturing body movement data in Sec. \ref{Capture_paradigm} and Sec. \ref{Capture_Methodology}. Emotion recognition datasets based on general body movements and gait patterns are reviewed in Sec. \ref{Posture_data} and Sec. \ref{gait_data}, respectively. Finally, Sec. \ref{Dataset_con} presents a detailed comparison of these datasets, highlighting their similarities and differences to inform and support future research efforts.

\subsection{Data Collection Paradigm}\label{Capture_paradigm}

\begin{figure}[tp]
    \centering
    \subfloat[Scene Settings for posture data collection]{\includegraphics[width=2in]{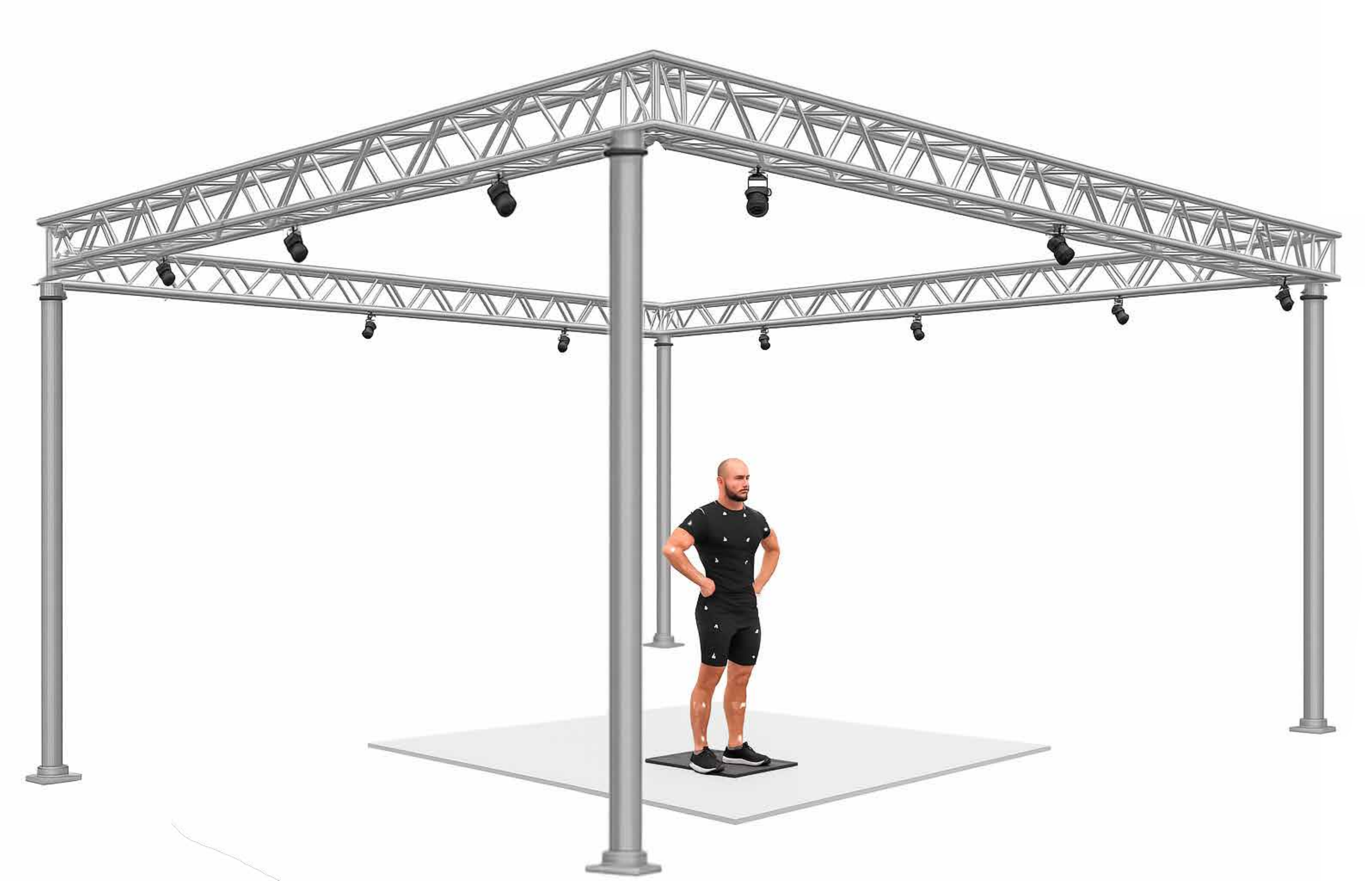}} \hspace{0.5cm}
    \subfloat[Scene Settings for gait data collection]{\includegraphics[width=2in]{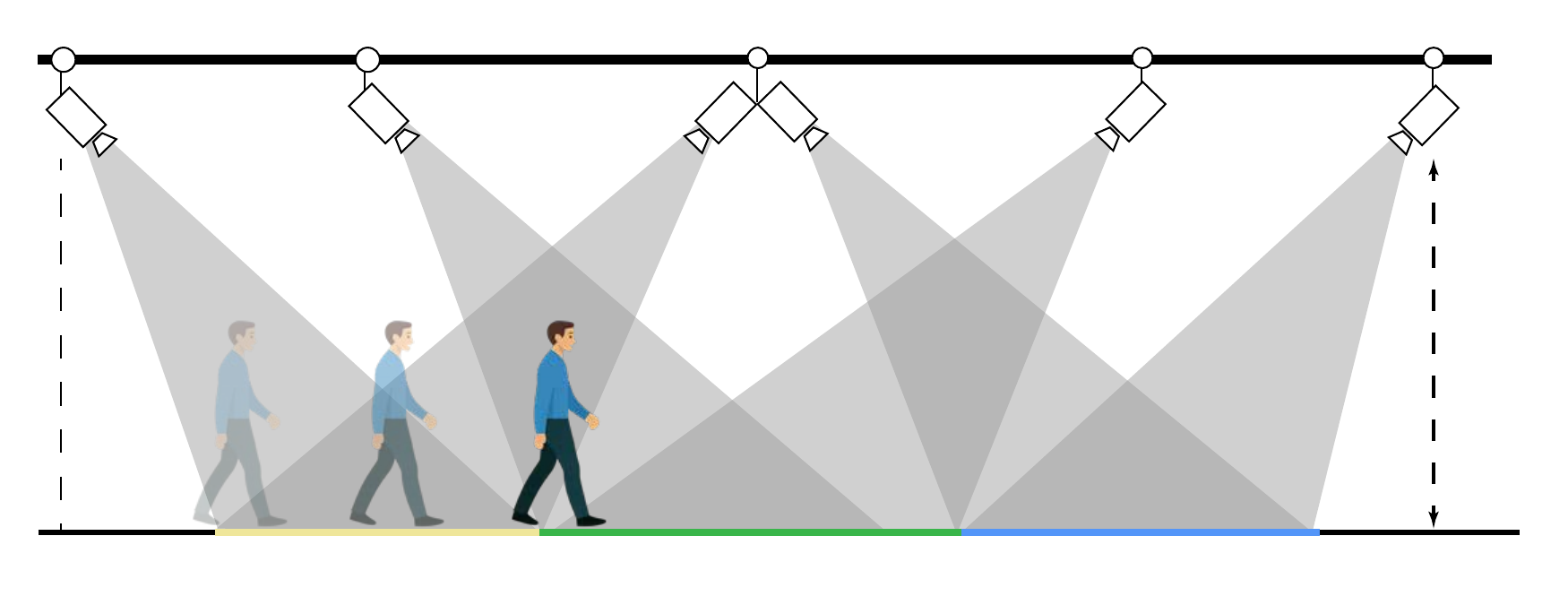}} \hspace{0.5cm}
    \caption{Comparison of different data collection scenarios}
    \label{Data_Collect}
\end{figure}

Since skeleton-based emotion recognition can be categorized into two types, there are correspondingly two data collection paradigms, as illustrated in Fig. \ref{Data_Collect}. Fig. \ref{Data_Collect}.(a) depicts the paradigm for collecting posture data, where actors typically perform various emotional expressions within a predefined area—usually less than one meter in size. In contrast, Fig. \ref{Data_Collect}.(b) illustrates the gait data collection paradigm, in which participants walk back and forth along a designated path, covering a larger activity area.


	

\subsection{Body Movement Capture Methodology}\label{Capture_Methodology}

\begin{figure}[tp]
	\centering   
	\centerline{\includegraphics[width=3.5in]{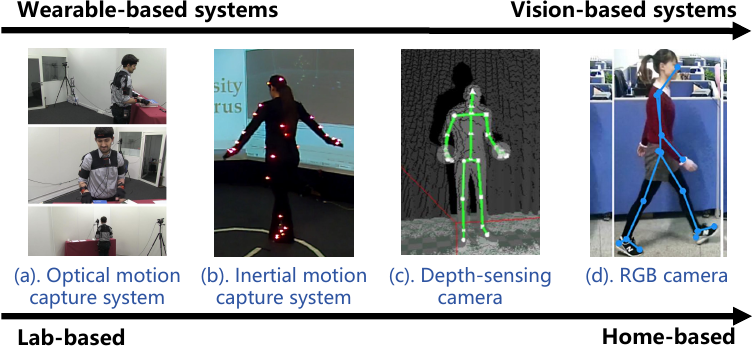}}
	\caption{Common methods for capturing 3D human skeleton data }
	\label{Data_collect_method}
\end{figure}

The methods for capturing 3D human skeleton data have evolved significantly over time, transitioning from professional, lab-based motion capture systems to more accessible, home-based solutions, as illustrated in Fig. \ref{Data_collect_method}. Initially, wearable-based systems such as optical motion capture systems (Fig. \ref{Data_collect_method}.(a)) were widely used in controlled laboratory environments. Subsequently, inertial motion capture systems (Fig. \ref{Data_collect_method}.(b)) emerged, utilizing wearable sensors such as Inertial Measurement Units (IMUs) to record body movements.

With advancements in sensor technology and computer vision, vision-based systems have gained increasing popularity due to their convenience and non-invasiveness. For example, depth-sensing cameras (Fig. \ref{Data_collect_method}.(c)) enable real-time skeleton tracking by capturing the 3D structural information of the body, while RGB cameras (Fig. \ref{Data_collect_method}.(d)) support image-based skeleton estimation through deep learning algorithms.

In the following sections, we provide a detailed overview of the characteristics and capabilities of each of these approaches.

\subsubsection{Optical Motion Capture System} 

In 2006, Ma et al. \cite{ma2006motion} employed an optical motion capture system to record three types of daily activities, annotating them with corresponding emotional labels. The DMCD dataset \cite{DMCD2021} was collected using a similar system to capture the movements of dancers. This system relies on reflective markers—typically spherical and coated with retro-reflective material—affixed to key anatomical landmarks on the subject’s body. Multiple infrared cameras positioned around the capture space emit light and detect the signals reflected from the markers. By triangulating the marker positions from different camera angles, the system can accurately reconstruct the 3D trajectories of joints. This method offers high spatial precision and is widely used in controlled laboratory environments for detailed motion analysis.

\subsubsection{Inertial Motion Capture System} 

The Emilya dataset \cite{fourati2016perception}, KDAE \cite{zhang2020kinematic}, and MEBED dataset \cite{volkova2014mpi} utilize inertial motion capture systems for data collection. These systems employ wearable IMU sensors, which are typically attached to major body segments such as the limbs, torso, and head. Each IMU contains accelerometers, gyroscopes, and, in some cases, magnetometers, enabling the measurement of linear acceleration, angular velocity, and orientation. By integrating these signals, the system can reconstruct the 3D motion of the human skeleton without the need for external cameras. IMU-based systems are portable and well-suited for capturing movement in unconstrained or real-world environments.

Additionally, Cui et al. \cite{cui2016emotion} utilized the built-in triaxial accelerometers in smartphones to monitor participants’ daily activities and classify their emotional states. In this study, two smartphones and one tablet were used: the smartphones were worn on the participant’s wrist and ankle to capture raw data from the accelerometer and gravity sensors, sampled at a rate of 5 Hz.

\subsubsection{Depth-sensing Camera} 

The release of the Kinect depth camera in 2011 \cite{zhang2012microsoft} significantly simplified the process of skeleton sequence acquisition. Since then, many researchers have adopted the Kinect series for data collection \cite{li2016emotion, 9718956}. Microsoft Kinect is a depth-sensing camera that integrates an RGB camera, an infrared (IR) emitter, and an IR depth sensor to capture both color and depth information. By projecting a structured infrared light pattern and analyzing its deformation, Kinect constructs a depth map of the scene. Using this depth data, built-in body tracking algorithms can identify human figures and estimate the 3D positions of joints in real time.

\subsubsection{RGB Camera} 

In 2017, breakthroughs in human pose estimation—particularly the development of real-time multi-person pose estimation methods such as OpenPose \cite{cao2017realtime}, HRNet \cite{sun2019deep}, and Vnect \cite{mehta2017vnect}—further advanced emotion recognition based on body movements. These techniques enabled the extraction of 2D or 3D skeletal joint coordinates directly from standard RGB video frames, eliminating the need for wearable sensors or depth cameras. The underlying algorithms typically employ Convolutional Neural Networks (CNNs) to detect body parts and estimate joint positions, followed by part affinity fields to associate joints with specific individuals in multi-person scenes. By 2019, researchers had begun leveraging skeleton sequences extracted from RGB videos for emotion recognition tasks \cite{randhavane2019liar, poyo2020role, ghaleb2021skeleton}.

\subsection{Posture-based Datasets}\label{Posture_data}

Table \ref{Dateset_List} summarizes datasets commonly used in skeleton-based emotion recognition research. Their distinctive features are discussed in detail in the following section.

\newcolumntype{W}[1]{>{\raggedright\arraybackslash}m{#1}}
\newcolumntype{C}[1]{>{\centering\arraybackslash}m{#1}}
	
	\begin{table}[tp]
		\centering
		\fontsize{9}{12}\selectfont
		\setlength{\tabcolsep}{1mm}
		\renewcommand\arraystretch{1.2}
		\caption{Overview of posture-based and gait-based datasets for skeleton-based emotion recognition}
		
		\begin{threeparttable}
			\begin{adjustbox}{max width=\textwidth}
				\begin{tabular}{W{1.4cm}C{2.6cm}C{2.6cm}C{1.1cm}C{1.3cm}C{1.1cm}C{1.8cm}C{1.2cm}}
					\toprule
					Type & Dataset & Acquisition Device & Subjects & Samples & Joints & Frame Rate & Emotions\\
					\midrule
					\multirow{7}{*}{\parbox[c]{1.4cm}{\centering Posture}}
					& \cellcolor{gray!15}Emilya~\cite{fourati2016perception}
					& \cellcolor{gray!15}Xsens MVN   & \cellcolor{gray!15}11
					& \cellcolor{gray!15}8206        & \cellcolor{gray!15}28
					& \cellcolor{gray!15}125 Hz      & \cellcolor{gray!15}8 \\
					
					& BML~\cite{ma2006motion}            & Optical MoCap & 30 & 4080 & 30 & 16 Hz  & 4 \\
					
					& \cellcolor{gray!15}MEBED~\cite{volkova2014mpi}
					& \cellcolor{gray!15}Xsens MVN     & \cellcolor{gray!15}8
					& \cellcolor{gray!15}1447          & \cellcolor{gray!15}23
					& \cellcolor{gray!15}120 Hz        & \cellcolor{gray!15}11 \\
					
					& KDAE~\cite{zhang2020kinematic}     & Noitom PN     & 22 & 1402 & 72 & 120 Hz & 7 \\
					
					& \cellcolor{gray!15}EGBM~\cite{sapinski2019multimodal}
					& \cellcolor{gray!15}Kinect V2     & \cellcolor{gray!15}16
					& \cellcolor{gray!15}560           & \cellcolor{gray!15}25
					& \cellcolor{gray!15}30 Hz         & \cellcolor{gray!15}7 \\
					
					& UCLIC~\cite{kleinsmith2006cross}   & Optical MoCap & 13 & 183  & 32 & --     & 4 \\
					
					& \cellcolor{gray!15}DMCD~\cite{DMCD2021}
					& \cellcolor{gray!15}Impulse X2    & \cellcolor{gray!15}6
					& \cellcolor{gray!15}108           & \cellcolor{gray!15}38
					& \cellcolor{gray!15}120 Hz        & \cellcolor{gray!15}12 \\
					
					\cmidrule(l){1-8}
					\multirow{5}{*}{\parbox[c]{1.4cm}{\centering Gait}}
					& E-Gait I~\cite{bhattacharya2020step,habibie2017recurrent}
					& Optical MoCap & -- & 1835 & 21 & 60 Hz & 4 \\
					
					& \cellcolor{gray!15}E-Gait II~\cite{bhattacharya2020step,habibie2017recurrent}
					& \cellcolor{gray!15}-- & \cellcolor{gray!15}--
					& \cellcolor{gray!15}342 & \cellcolor{gray!15}16
					& \cellcolor{gray!15}--  & \cellcolor{gray!15}4 \\
					
					& BME I~\cite{Hicheur2013}           & Vicon V8 & 8 & 200 & 12 & 120 Hz & 5 \\
					
					& \cellcolor{gray!15}BME II~\cite{Hicheur2013}
					& \cellcolor{gray!15}Vicon V8       & \cellcolor{gray!15}5
					& \cellcolor{gray!15}75             & \cellcolor{gray!15}12
					& \cellcolor{gray!15}120 Hz         & \cellcolor{gray!15}5 \\
					
					& EMOGAIT~\cite{sheng2021multi}      & RGB Camera & 60 & 1440 & 16 & -- & 4 \\
					\bottomrule
				\end{tabular}
			\end{adjustbox}
		\end{threeparttable}
		
		\label{Dateset_List}
	\end{table}

\textbf{Emilya Dataset.} The EMotional body expression In daILY Actions dataset (Emilya) \cite{fourati2014emilya, fourati2016perception} comprises a total of 8,206 samples. Eleven actors (six females and five males) performed eight distinct emotions within the context of seven everyday actions. These actions were carefully selected to involve full-body movements, particularly emphasizing the upper body and arm motions, in order to ensure diverse emotional expressions. The seven daily activities include sitting down, walking, walking while carrying an object, moving books across a table with both hands, knocking on a door, lifting an object, and throwing an object.

In the walking tasks, actors were instructed to walk back and forth along the longer side of a room, with two distinct variations: normal walking and walking while carrying an object. Each sample was labeled according to one of eight emotional categories: Neutral, Joy, Anger, Panic, Fear, Anxiety, Sadness, and Shame.

All data were captured using the Xsens MVN motion capture system, which records 3D skeleton sequences at a frame rate of 120 Hz. Each skeleton sequence consists of 28 joints.

\textbf{BML Dataset.}
The BML Dataset \cite{ma2006motion} comprises 4,080 body movement sequences. Thirty actors (15 females and 15 males) performed four distinct actions—walking, knocking, lifting, and throwing—each interleaved with walking segments, under four different emotional states: angry, happy, neutral, and sad. All motion data were captured using the Falcon Analog optical motion capture system, which provides 3D positional data for 16 joints. 

\textbf{MEBED Dataset.}
The MPI Emotional Body Expression Database (MEBED) consists of 1,477 samples. Eight actors (four males and four females), each around 25 years old, participated in the data collection. The participants were asked to perform emotional expressions across four scenario types, each containing ten distinct emotion categories. The four scenario types are: Solitary non-verbal emotional scenarios, Communicative non-verbal emotional scenarios, Short sentences without direct speech, and Short sentences with direct speech. The ten emotions include: amusement, joy, pride, relief, surprise, anger, disgust, fear, sadness, and shame, with an additional neutral emotion category included.

All motion data were recorded using the Xsens MVN motion capture system at a sampling rate of 120 Hz, capturing 3D skeletal sequences composed of 28 joints. During the recording sessions, all participants were seated on stools. Consequently, the ten lower-body joints are typically excluded during analysis.

MEBED includes dual-label annotations: self-reported emotions from the performers and third-party annotations from external observers viewing each motion sequence. It is worth noting that the dataset exhibits class imbalance, with significant variation in the number of samples per emotion category.

\textbf{KDAE Dataset.}
The Kinematic Dataset of Actors Expressing Emotions (KDAE) \cite{zhang2020kinematic} contains a total of 1,402 samples. Twenty-two semi-professional actors (11 females and 11 males) were asked to perform seven distinct emotions—happiness, sadness, neutral, anger, disgust, fear, and surprise—across 70 everyday event scenarios (10 scenarios per emotion). The actors performed the emotional expressions on a 1m $\times$ 1m square stage, positioned 0.5 meters away from a wall.

All motion data were recorded using a portable wireless motion capture system capable of tracking 72 body markers at a frame rate of 125 Hz. Notably, nearly half of these 72 markers correspond to the hands. However, as this work focuses specifically on full-body emotional expression, the hand markers were excluded from the analysis. Only 24 key body markers were retained.

\textbf{EGBM Dataset.}
The Emotional Gestures and Body Movements Corpus (EGBM) \cite{sapinski2019multimodal} consists of 560 samples. The dataset features performances by 16 professional Polish actors (8 females and 8 males), who expressed seven distinct emotions—happiness, sadness, neutral, anger, disgust, fear, and surprise—without any predefined instructions or cues.

All motion data were captured using a Kinect V2 camera at a frame rate of 30 Hz. Each emotion category contains 80 samples, and each pose sequence provides the 3D positions of 25 joints.

\textbf{UCLIC Dataset.}
The UCLIC dataset \cite{kleinsmith2006cross} comprises 108 samples. It features performances from 13 actors, including 11 Japanese, 1 Sri Lankan, and 1 American. The actors expressed four basic emotions—anger, fear, happiness, and sadness—freely and based on their own interpretations, without any imposed constraints. 

All motion data were collected using a MoCap system, with each pose sequence providing the 3D positions of 32 joints.

\textbf{DMCD Dataset.}
The Dance Motion Capture Database (DMCD) \cite{DMCD2021} consists of 108 samples. The dataset features performances by six female dancers with diverse backgrounds in gymnastics, ballet, theatrical dance, and other styles. Each participant performed a dance sequence, with each sequence associated with a specific emotional state.

In total, 12 emotions were expressed: excited, happy, pleased, satisfied, relaxed, tired, bored, sad, miserable, annoyed, angry, and afraid. All motion sequences were captured using the PhaseSpace Impulse X2 MoCap System at a frame rate of 120 Hz. Each pose sequence contains the 3D positions of 38 joints.


\subsection{Gait-based Datasets}\label{gait_data}

\textbf{E-Gait Dataset.}
The Emotion-Gait (E-Gait) dataset \cite{bhattacharya2020step} comprises 2,177 real-world gait sequences, each annotated with one of four emotion categories: happy, sad, neutral, or angry. The dataset is divided into two subsets:
\begin{itemize}
	\item Subset A: Contains 342 gait sequences sourced from datasets including BML \cite{ma2006motion}, ICT \cite{narang2017motion}, CMU-MOCAP \cite{cmumocap}, and Human3.6M \cite{6682899}.
	\item Subset B: Comprises 1,835 gait sequences extracted from the Edinburgh Locomotion Mocap Database (ELMD) \cite{habibie2017recurrent}.
\end{itemize}

All motion data were captured using a motion capture (MoCap) system at a frame rate of 60 Hz, recording 3D positions of 21 joints. To ensure consistency across both subsets, the authors standardized the skeletal data to include 16 joints. Each gait sequence in the E-Gait dataset was independently annotated by 10 individuals. The final emotion label for each sequence was determined through majority voting among the annotators.

\textbf{BME Dataset.}
The Body Motion-Emotion dataset (BME), dataset \cite{Hicheur2013} dataset involves two experiments.
\begin{itemize}
	\item Emotional Gaits: 8 professional actors (4 males, 4 females) performed five emotions—neutral, joy, anger, sadness, and fear—while walking. Each emotion was recorded in at least 5 trials, resulting in 200 total samples. Actors were instructed to embody emotions (e.g., "walking in a dark, dangerous room" for fear) to standardize expressions.
	\item Control Gaits: 5 naive male subjects walked at three speeds—slow, normal, and fast —with 5 repetitions per speed, yielding 75 total samples. This subset isolated speed-related kinematic variations for comparison.
\end{itemize}

All motion data were captured using a Vicon V8 optoelectronic motion capture system with 24 cameras (16 in Experiment 2) at a frame rate of 120 Hz. The system tracked 3D positions of markers on 12 body segments, including joints like shoulders, elbows, hips, and knees. 

\textbf{EMOGAIT Dataset.}
The EMOGAIT dataset \cite{sheng2021multi} consists of 1,440 real-world gait sequences collected from 60 subjects (33 females and 27 males). Each sequence is annotated with one of four emotion labels: happy, sad, neutral, or angry. All data were captured using standard RGB cameras, and 2D human skeletons were extracted from the videos using a pose estimation algorithm. Each skeleton sequence contains 16 joints.

\subsection{Dataset Comparison and Analysis}\label{Dataset_con}

\newcolumntype{L}[1]{>{\raggedright\arraybackslash}p{#1}}
\newcolumntype{C}[1]{>{\centering\arraybackslash}p{#1}}

\begin{table}[tp]
	\centering
	\fontsize{9}{12}\selectfont
	\renewcommand\arraystretch{1.2}
	\caption{Comparison of emotion induction methods and performance instructions across different datasets}
	\label{Dateset_com}
	
	\begin{adjustbox}{max width=\linewidth}
		
		
		\begin{tabular}{L{2cm} C{2.5cm} L{4cm} L{4.5cm}}
			\toprule
			Dataset & Performer & Emotion Induction Method & Performance Instructions \\
			\midrule
			
			\rowcolor{gray!12}
			Emilya \cite{fourati2016perception}  
			& College students  
			& Pre-selected emotional scenarios  
			& Perform specific movements while expressing the given emotions \\
			
			BML \cite{ma2006motion}  
			& College students  
			& Pre-selected emotional scenarios  
			& Perform specific movements while expressing the given emotions \\
			
			\rowcolor{gray!12}
			MEBED \cite{volkova2014mpi}  
			& Amateur actors  
			& Pre-selected emotional scenarios  
			& Sit on a chair and read aloud with matching gestures \\
			
			KDAE \cite{zhang2020kinematic}  
			& College students  
			& Pre-selected emotional scenarios  
			& Performers act freely based on the given scenarios \\
			
			\rowcolor{gray!12}
			EGBM \cite{sapinski2019multimodal}  
			& Professional actors  
			& Pre-selected emotional scenarios  
			& Performers act freely based on the given scenarios \\
			
			UCLIC \cite{kleinsmith2006cross}  
			& Unknown  
			& None  
			& Perform emotional postures freely with no specific constraints \\
			
			\rowcolor{gray!12}
			EMOGAIT \cite{sheng2021multi}  
			& College students  
			& Watching various emotional film clips  
			& Walk back and forth after watching the clips \\
			\bottomrule
		\end{tabular}
	\end{adjustbox}
\end{table}

Table \ref{Dateset_com} provides a comparative overview of emotion induction methods and performance instructions across several widely used datasets. Most datasets—such as Emilya, BML, MEBED, KDAE, and EGBM—utilize pre-selected emotionally charged scenarios to elicit specific emotional expressions from participants. While datasets like Emilya and BML offer explicit movement instructions to ensure consistency, others such as KDAE and EGBM adopt a freer performance style, allowing participants to express emotions more naturally. The level of performer expertise also varies across datasets, ranging from college students and amateur actors to trained professionals, potentially influencing both the expressiveness and reliability of the recorded data.

The EMOGAIT dataset is notable for its use of emotionally evocative film clips as an induction method, aiming to capture changes in gait following the emotional priming. In contrast, the UCLIC dataset is distinctive in its lack of both a defined emotion induction protocol and specific performance instructions, relying instead on participants’ spontaneous emotional postures. These differences reflect the varied design philosophies and intended applications of each dataset, ultimately affecting the naturalness, consistency, and emotional richness of the captured body movement data.

\section{Body Movement Based Emotion Recognition}\label{sec_4}

\begin{figure}[tp]
    \centering
    \subfloat[Traditional approaches]{\includegraphics[width=2.5in]{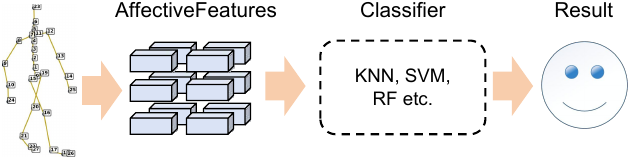}} \hspace{0.5cm}
    \subfloat[Feat2Net approaches]{\includegraphics[width=2.5in]{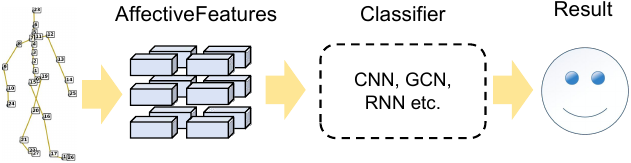}} \\
    \subfloat[FeatFusionNet approaches]{\includegraphics[width=2.5in]{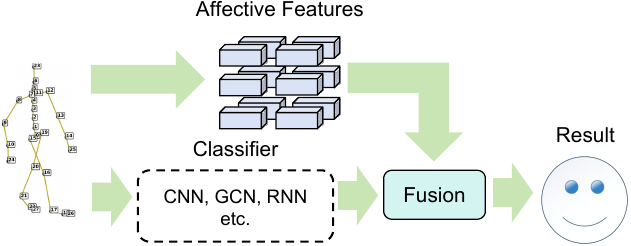}} \hspace{0.5cm}
    \subfloat[End2EndNet approaches]{\includegraphics[width=2.5in]{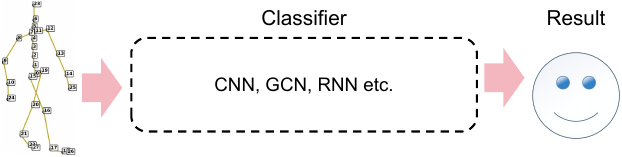}}
    \caption{Comparison of different technical approaches}
    \label{Method_diff}
\end{figure}

This section provides a comprehensive overview of research advances in skeleton-based emotion recognition. From the perspectives of data characteristics and technical development, existing methods can be broadly categorized into two main groups: \textbf{posture-based} (in Sec. \ref{posture_based_method}) and \textbf{gait-based} (in Sec. \ref{gait_based_method}) approaches. Within each group, methods are further classified into four representative categories based on their technical design:
\begin{itemize}
	\item \textbf{Traditional approaches} (see Fig. \ref{Method_diff}.(a)), which involve extracting handcrafted affective features (e.g., joint angles, body symmetry, velocity) from skeleton data, followed by machine learning classifiers for emotion recognition;
	
	\item \textbf{Feat2Net approaches} (see Fig. \ref{Method_diff}.(b)), which extract handcrafted features from skeleton sequences and employ neural networks solely for classification;
	
	\item \textbf{FeatFusionNet approaches} (see Fig. \ref{Method_diff}.(c)), which extract handcrafted features and integrate them into the training process of deep learning models to enhance discriminative performance;
	
	\item \textbf{End2EndNet approaches} (see Fig. \ref{Method_diff}.(d)), which directly learn body movement representations and perform emotion recognition from raw skeleton data without relying on manually designed features.
\end{itemize} 

The following subsections are structured based on this taxonomy, offering a detailed analysis of the evolution, key representative works.

\subsection{Posture-based Emotion Recognition}\label{posture_based_method}

\subsubsection{Traditional Approaches} 

Machine learning–based methods have long served as a foundational component in the field of emotion recognition from body movements. Table \ref{Posture_ML_List} summarizes a selection of representative studies from recent years that focus on automatic emotion recognition using human motion data. To facilitate a clearer understanding and comparison of posture-based emotion recognition approaches, this section is organized according to the data acquisition techniques employed to capture posture information. Specifically, we categorize existing works based on whether the posture data is collected using motion capture systems, depth sensors, or RGB cameras.

\newcolumntype{L}[1]{>{\raggedright\arraybackslash}p{#1}}
\newcolumntype{C}[1]{>{\centering\arraybackslash}p{#1}}

\begin{table*}[tp]
	\centering
	\fontsize{9}{12}\selectfont
	\setlength{\tabcolsep}{1.5mm}   
	\renewcommand\arraystretch{1.2}
	\caption{Emotion recognition from posture using machine learning methods}
	\label{Posture_ML_List}
	
	\begin{adjustbox}{max width=\textwidth}
		
		
		\begin{threeparttable}
			\begin{tabular}{@{}L{1.9cm} L{2.2cm} L{5.4cm} C{2.4cm} C{1.5cm} L{2.8cm}@{}}
				\toprule
				Study & Dataset & Feature & Classifier & Protocol & Accuracy \\
				\midrule
				
				\rowcolor{gray!12}
				Kapur \textit{et al.} \cite{kapur2005gesture} &
				5 subjects, 500 samples &
				Mean/SD of position, velocity, acceleration &
				LR / NB / DT / MLP / SVM &
				10-fold, LOSO &
				91.8 \%; 84.6 \%$\pm$12.1 \% \\
				
				Bernhardt \textit{et al.} \cite{bernhardt2007detecting} &
				BML &
				Max hand distance/speed/acceleration/jerk &
				SVM &
				LOSO &
				81.1 \% (Sensitivity) \\
				
				\rowcolor{gray!12}
				Fourati \textit{et al.} \cite{fourati2016perception} &
				Emilya &
				Power, Fluidity, Speed, Quantity/Regularity, Body openness, Leaning, Straightness &
				MLR &
				3-fold &
				36–48 \% \\
				
				Fourati \textit{et al.} \cite{Fourati2015} &
				Emilya &
				110 expressive body cues (multi-level notation) &
				RF &
				OOB &
				67.9–84.8 \% \\
				
				\rowcolor{gray!12}
				Fourati \textit{et al.} \cite{fourati2015relevant} &
				Emilya &
				114 expressive body cues (multi-level notation) &
				RF &
				OOB &
				78.29–91.53 \% \\
				
				Fourati \textit{et al.} \cite{fourati2019contribution} &
				Emilya &
				114 cues; 80 position; 80 kinetic-energy feats &
				RF / SVM &
				3-fold &
				73.93–74.47 \% (F1) \\
				
				\rowcolor{gray!12}
				Crenn \textit{et al.} \cite{Crenn2016} &
				BML / UCLIC / SIGGRAPH &
				68 low-level geometric, motion, Fourier feats &
				SVM &
				10-fold &
				BML 57 \% / UCLIC 78 \% / SIGGRAPH 93 \% \\
				
				Crenn \textit{et al.} \cite{Crenn2017} &
				BML / MEBED / UCLIC / SIGGRAPH &
				Spectral amplitude differences (neutral vs expressive) &
				SVM / RF / KNN &
				10-fold &
				57 \% / 67 \% / 83 \% / 98 \% \\
				
				\rowcolor{gray!12}
				Crenn \textit{et al.} \cite{crenn2020generic} &
				Emilya / MEBED / UCLIC / SIGGRAPH &
				Posture, Temporal, Residue features &
				SVM / RF / KNN &
				10-fold &
				82.2 \% / 78.6 \% / 74 \% / 98.8 \% \\
				
				Saha \textit{et al.} \cite{saha2014study} &
				10 subjects &
				Hand–spine dist., max acc., various joint angles &
				DT / KNN / SVM / NN &
				N/A &
				76.63–90.83 \% \\
				
				\rowcolor{gray!12}
				Piana \textit{et al.} \cite{piana2016adaptive} &
				Qualisys (12 ppl, 310 seg.) / Kinect (579 seg.) &
				Holistic (kinetic, contraction, symmetry…) and local features &
				SVM &
				Hold-out, LOSO &
				Qualisys 62.3 $\pm$ 25.3 \%, Kinect 68.5 $\pm$ 18.5 \%;  
				Qualisys 54.2 $\pm$ 29.1 \%, Kinect 61.6 $\pm$ 23.1 \% \\ 
				\bottomrule
			\end{tabular}
			
			\begin{tablenotes}
				\footnotesize
				\item LR: Logistic Regression; NB: Naive Bayes; DT: Decision Tree; MLP: Multilayer Perceptron; SVM: Support Vector Machine; MLR: Multinomial Logistic Regression; RF: Random Forest; KNN: K-Nearest Neighbor; NN: Neural Network; LOSO: Leave-One-Subject-Out; OOB: Out-of-Bag.
			\end{tablenotes}
		\end{threeparttable}
	\end{adjustbox}
\end{table*}

\textbf{Motion Capture Systems.}
Several studies have focused on extracting low-level kinematic features from skeleton or motion data to recognize affective states. Kapur et al. \cite{kapur2005gesture} laid the foundation by extracting the mean and standard deviation of position, velocity, and acceleration as input features for training a machine learning model to classify human emotions. Bernhardt et al. \cite{bernhardt2007detecting} advanced this line of work by segmenting complex movements into primitives, analyzing key dynamic features, and applying normalization techniques to reduce individual movement biases. Extending these efforts, Samadani et al. \cite{Samadani2014} proposed a hybrid generative-discriminative approach for affective movement recognition that accounts for kinematic variation, interpersonal differences, and stochastic noise. Their method employs Hidden Markov Models (HMMs) to capture the temporal dynamics of affective movements and generates Fisher Score representations to encode both kinematic and dynamic characteristics.

A notable contribution to emotion recognition from body movement comes from the work of Fourati et al., who developed a multi-level framework for systematically describing and analyzing emotional body behaviors. In their 2014 study \cite{fourati2014collection}, the authors introduced the Multi-Level Body Notation System (MLBNS) along with a corresponding emotional body behavior dataset. This system employed a movement-quality-based coding scheme with three descriptive levels, enabling a structured characterization of body movement features. In addition to the coding framework, they recorded and validated a new dataset to support further analysis. To simplify the perceptual rating task, a follow-up study \cite{fourati2016perception} selected a subset of key variables from MLBNS. Building on this foundation, Fourati et al. \cite{Fourati2015} expanded the framework by extracting 110 emotion-related features across three hierarchical levels: anatomical description, directional description, and posture/movement dynamics. These features included 38 body part descriptors, 30 inter-limb relationship measures, and 42 local limb movement descriptors. While both \cite{Fourati2015} and \cite{fourati2015relevant} pursued similar objectives, the latter provided a more detailed analysis of the relative importance of individual features for distinguishing emotional categories. Further extending the MLBNS framework, the authors later introduced four additional feature groups—temporal patterns, slope-based descriptors, peak dynamics, global motion statistics, and temporal regularity measures—which led to notable improvements in classification accuracy \cite{fourati2019contribution}.

Another influential line of research has been led by Crenn et al., who systematically investigated emotion recognition through detailed analysis of skeletal motion features. In their early work, Crenn et al. \cite{Crenn2016} proposed a comprehensive feature extraction framework that integrates geometric relationships, motion correlations, and frequency-based periodicity. The extracted features included inter-joint distances, triangle areas formed by specific joints, joint angles, and the velocity and acceleration of various joints. Experimental results showed that these features were strongly correlated with emotional expression. Building on this foundation, Crenn et al. \cite{Crenn2017} introduced the concept of a neutral pose, suggesting that emotional states could be inferred by analyzing the residual differences between observed skeletal data and this neutral reference. This concept was further refined in a 2020 study \cite{crenn2020generic}, where a cost function optimization strategy was employed to more effectively synthesize neutral motion, thereby improving emotion recognition performance. In subsequent research, the motion-emotion feature framework proposed by Crenn has been widely adopted and extended by others, highlighting its lasting impact on the field.

Emotion recognition based on full-body movement data in gaming scenarios has emerged as a significant and increasingly influential research direction. Kleinsmith et al. \cite{kleinsmith2011automatic} investigated the recognition of non-basic emotional states within video game contexts. They collected posture data from players engaged in Nintendo Wii sports games, established ground-truth labels through human observer evaluations, and developed automatic recognition models based on the captured movements. Expanding on this idea, Savva et al. \cite{savva2012continuous} explored whether players’ full-body movements in gaming environments could reflect their aesthetic and emotional experiences. Using similar Nintendo sports games, their study demonstrated that the automatic system achieved an emotion recognition accuracy comparable to that of human raters. Further advancing this line of research, Garber et al. \cite{garber2012using} collected motion capture data during gameplay and extracted features such as body symmetry, head displacement, and posture openness to infer players’ emotional states.

\textbf{Depth Sensors.}
Saha et al. \cite{saha2014study} conducted research on gesture-based emotion recognition using a Kinect sensor. They selected eleven joint coordinates from the upper body and hands to extract nine features related to joint distances, accelerations, and angles, which were used to recognize five basic emotions: anger, fear, happiness, sadness, and relaxation. The results showed that an ensemble decision tree achieved the highest average classification accuracy of 90.83\%. Piana et al. \cite{piana2016adaptive} processed motion capture data to extract features such as joint energy, movement direction, posture, and body symmetry. They constructed adaptive representations through dictionary learning and employed a linear support vector machine for emotion classification. Ahmed et al. \cite{ahmed2019emotion} computed a comprehensive set of body movement features categorized into ten groups. In the first stage, they applied analysis of variance (ANOVA) and multivariate analysis of variance (MANOVA) to remove irrelevant features and distribute the remaining ones across groups. In the second stage, a binary chromosome–based genetic algorithm was used to select an optimal feature subset for maximizing emotion recognition performance. Finally, score- and rank-level fusion techniques were applied to further enhance classification accuracy.

\textbf{RGB Camera.} 
Glowinski et al. \cite{Glowinski2008} proposed a method for automatic emotion recognition by analyzing upper-body gestures, specifically head and hand movements. Using 40 emotional clips—representing anger, joy, relief, and sadness—performed by professional actors from the GEMEP database \cite{GEMEP2025}, the researchers captured video recordings with a dual-camera setup and extracted dynamic features such as motion energy (sum of velocity magnitudes) and spatial extent (perimeter of the enclosing triangle). In a follow-up study, Glowinski et al. \cite{glowinski2011toward} extracted additional kinematic features from head and hand trajectories, including energy, spatial range, smoothness, and symmetry. They applied Principal Component Analysis (PCA) to reduce the feature dimensionality to a four-dimensional representation, which effectively grouped emotions along the axes of valence (positive vs. negative) and arousal (high vs. low).

\subsubsection{Feat2Net Approaches}
In \cite{avola2020deep}, a multi-branch deep learning architecture is proposed, consisting of a local branch based on stacked Long Short-Term Memory (LSTM) units and a global branch based on a Multi-Layer Perceptron (MLP). The local branch processes temporal local features through stacked LSTM layers to achieve higher levels of abstraction and capture fine-grained action details. In parallel, the global branch utilizes a MLP to process temporal global features and extract salient patterns. The outputs from both branches are concatenated and passed through a fully connected layer followed by a softmax function for emotion classification.

In a more recent study, O{\u{g}}uz et al.~\cite{ouguz2024emotion} investigated a wide range of time-domain, frequency-domain, and statistical features, applying feature selection techniques to identify four salient features per frame. These selected features were then aggregated into a feature matrix used for subsequent emotion recognition.

Wang et al.\cite{wang2023emotion} proposed a multi-scale feature selection algorithm grounded in a pseudo-energy model and introduced a multi-scale spatiotemporal network to decode the complex relationship between emotional states and full-body movements. This approach effectively captures both long-term postural variations and short-term dynamics, thereby enhancing emotion recognition performance. Further extending this line of research, Wang et al.\cite{wang2025affective} explored joint energy features by constructing a body expression energy model and designing a multi-input symmetric positive definite matrix network. This framework facilitates the extraction of interpretable spatiotemporal features, contributing to both the robustness and interpretability of emotion classification.

\subsubsection{End2EndNet Approaches}
In recent years, with the remarkable success of neural networks in the broader field of artificial intelligence, deep learning models have become the dominant paradigm for emotion recognition from 3D skeletal motion sequences. In this section, we summarize existing methods based on differences in their network architectures.

\textbf{RNN-based models.}
Sapiński et al.\cite{sapinski2019emotion} collected body movement data from professional actors using Kinect V2 and employed LSTM networks to classify emotions based on skeletal sequences. Zhang et al.\cite{zhang2021emotion} proposed an Attention-based Stacked LSTM (AS-LSTM) model for emotion recognition from whole-body movements in virtual reality (VR) environments. By integrating an attention mechanism into the traditional LSTM framework, the model assigns varying weights to joint point sequences across motion frames, enabling it to focus on key joints while suppressing redundant information. This enhancement improves both the learning capacity of the network and the overall recognition accuracy.

\textbf{CNN-based models.}
Karumuri et al.\cite{karumuri2019motions} were among the first to propose encoding skeletal information into image representations. They converted 3D joint coordinates into 8-bit RGB images using four encoding schemes: coarse position, fine position, logical position, and logical velocity. Two convolutional neural network (CNN) architectures were then designed for training—Single-Input Architecture (SIA-CNN) and Multi-Input Architecture (MIA-CNN). Cui et al.\cite{mingming2020emotion} drew on findings from body language literature to associate specific postures with emotions. They employed the Convolutional Pose Machine (CPM) algorithm to extract human keypoint coordinates, generated simplified line diagrams via clustering, and used a CNN combined with a Softmax layer for emotion classification. Beyan et al.~\cite{Beyan2023TAC} proposed a dual-branch CNN architecture that processes coarse-grained (e.g., 4s) and fine-grained (e.g., 1s) temporal features in parallel, leveraging logical position image representations and data augmentation techniques to improve classification performance.

\textbf{GCN-based models.}
Shen et al.\cite{shen2019emotion} employed a Temporal Segment Network (TSN) to extract RGB features and used ST-GCN for skeletal features. These features were unified through a residual encoder and classified via a residual fully connected network. Ghaleb et al.\cite{ghaleb2021skeleton} developed an emotion recognition framework based on Spatial-Temporal Graph Convolutional Networks (ST-GCN), enhanced with a spatial attention mechanism to highlight the relationship between joint spatial patterns and emotional states. Building on the ST-GCN architecture, Shi et al.\cite{Shi2021skeleton_STGCN} introduced a self-attention mechanism that dynamically adjusts the skeletal connectivity structure, demonstrating the effectiveness of adaptive topologies for capturing emotional cues. Shirian et al.\cite{shirian2021dynamic} proposed the Learnable Graph Inception Network (L-GrIN), which incorporates non-linear spectral graph convolution, a graph inception layer, learnable adjacency, and a learnable pooling function. The model jointly learns the graph structure and performs emotion classification by optimizing a composite loss function that combines classification and graph structure losses.

\subsubsection{Pre-training Approaches}
With the rapid advancement of large-scale models, researchers have increasingly recognized the value of pretrained models in providing transferable prior knowledge across diverse domains, thereby significantly enhancing task performance. Consequently, there has been growing interest in incorporating pretraining strategies into skeleton-based emotion recognition.

Paiva et al. \cite{paiva2025skelett} adopted a two-stage approach: self-supervised pretraining on large-scale unlabeled skeleton datasets (MPOSE2021 and Panoptic Studio) using a Masked Autoencoder (MAE) to learn spatiotemporal dependencies, followed by fine-tuning on the BoLD dataset using an MLP classifier. In \cite{Lu_Chen_Liang_Tan_Zeng_Hu_2025}, the authors proposed the Emotion-Action Interpreter based on Large Language Models (EAI-LLM). This model first extracts skeleton features using a Graph Convolutional Network (GCN), and then maps these features into a semantic space through components such as a multi-granularity skeletal annotator and a unified skeletal labeling module. By leveraging the reasoning capabilities of large language models, EAI-LLM not only performs emotion recognition but also generates interpretable textual explanations.

\subsubsection{Evaluation of Methods on Public Datasets}

To facilitate a clear comparison of different approaches, we summarize the performance of representative methods on three widely used public datasets—EGBM, KDAE, and Emilya—as shown in Tables \ref{Dataset_EGBM_List}–\ref{Dataset_Emilya_List}.

On the EGBM dataset, early RNN-based methods achieved moderate accuracy (69\% and 74\%, respectively). In contrast, more recent approaches that incorporate handcrafted features with fully connected networks (FC), such as those proposed by Wang et al. \cite{wang2025affective, wang2023emotion}, report significantly higher performance, exceeding 95\%. A similar pattern is observed on the KDAE dataset. While GCN-based models \cite{ghaleb2021skeleton} yielded modest results (65\%), methods combining handcrafted features with shallow classifiers or neural networks demonstrated strong performance. Notably, O\u{g}uz et al. \cite{ouguz2024emotion} achieved an impressive 99.99\% accuracy under a hold-out protocol. On the Emilya dataset, both traditional and deep learning models performed well; CNN-based methods \cite{Beyan2023TAC} and handcrafted feature–driven fully connected networks \cite{wang2023emotion} both surpassed 94\% accuracy.

Overall, while end-to-end models generally yield superior performance, handcrafted feature–based approaches remain highly competitive—particularly in scenarios with limited data availability.

\begin{table}[tp]  
	\setlength{\tabcolsep}{1mm}
	\renewcommand\arraystretch{1.2}
	\centering 
	\fontsize{9}{12}\selectfont
	\caption{Method comparison on EGBM dataset}  
	\label{Dataset_EGBM_List}
	\begin{threeparttable}  
		\begin{tabular}{@{}llll@{}}
			\toprule
			Study                                                                           & Backbone           & Protocol  & Accuracy        \\ \midrule
			\rowcolor{gray!12}
			Sapi{\'n}ski et al.\cite{sapinski2019emotion} & RNN            & 10 - fold & 69.00\%            \\
			AS-LSTM \cite{zhang2021emotion}                           & RNN                & 10 - fold & 74.00\%            \\
			
			\rowcolor{gray!12}
			Wang et al. \cite{wang2025affective}                          & Manual features+FC & 10 - fold & 97.43\%         \\
			Wang et al. \cite{wang2023emotion}                             & Manual features+FC & 10 - fold & 95.55\% \\
			
			\rowcolor{gray!12}
			EAI-LLM \cite{Lu_Chen_Liang_Tan_Zeng_Hu_2025}           & LLMs               & Hold-out  & 66.97\%         \\ \bottomrule
		\end{tabular}
	\end{threeparttable} 
\end{table}

\begin{table}[tp]  
	\setlength{\tabcolsep}{1mm}
	\renewcommand\arraystretch{1.2}
	\centering  
	\fontsize{9}{12}\selectfont
	\caption{Method comparison on KDAE dataset}  
	\label{Dataset_KDAE_List}
	\begin{threeparttable}  
		\begin{tabular}{@{}llll@{}}
			\toprule
			Study                                                                 & Backbone           & Protocol  & Accuracy        \\ \midrule
			\rowcolor{gray!12}
			Ghaleb et al. \cite{ghaleb2021skeleton}               & GCN                & 10 - fold & 65.00\%            \\
			
			Wang et al. \cite{wang2025affective}                 & Manual features+FC & 10 - fold & 96.67\%         \\
			
			\rowcolor{gray!12}
			Wang et al. \cite{wang2023emotion}                   & Manual features+FC & 10 - fold & 95.60\%  \\
			O{\u{g}}uz et al. \cite{ouguz2024emotion}                  & Manual features+NN & Hold-out & 99.99\% \\
			
			\rowcolor{gray!12}
			EAI-LLM\cite{Lu_Chen_Liang_Tan_Zeng_Hu_2025} & LLMs               & Hold-out  & 71.17\%         \\ \bottomrule
		\end{tabular}
	\end{threeparttable} 
\end{table}

\begin{table}[tp]  
	\setlength{\tabcolsep}{1mm}
	\renewcommand\arraystretch{1.2}
	\centering  
	\fontsize{9}{12}\selectfont
	\caption{Method comparison on Emilya dataset}  
	\label{Dataset_Emilya_List}
	\begin{threeparttable}  
		\begin{tabular}{@{}llll@{}}
			\toprule
			Study                                                                 & Backbone           & Protocol  & Accuracy        \\ \midrule
			
			\rowcolor{gray!12}
			Beyan et al. \cite{Beyan2023TAC}                & CNN                & 5 - fold  & 96.59\%         \\
			Wang et al. \cite{wang2023emotion}                   & Manual features+FC & 10 - fold & 94.42\%  \\
			
			\rowcolor{gray!12}
			EAI-LLM \cite{Lu_Chen_Liang_Tan_Zeng_Hu_2025} & LLMs               & Hold-out  & 85.44\%         \\ \bottomrule
		\end{tabular}
	\end{threeparttable} 
\end{table}

\subsection{Gait-based Emotion Recognition}\label{gait_based_method}

\subsubsection{Traditional Approaches}

Machine learning–based methods have consistently served as a cornerstone in the field of emotion recognition from gait. Table \ref{Gait_DL_List} summarizes a selection of representative studies in recent years that focus on automatic emotion recognition based on human gait. To facilitate a clearer understanding and comparison of gait-based emotion recognition approaches, this section is organized according to the data acquisition methods used for capturing gait information. Specifically, we categorize existing works based on whether gait data is obtained through motion capture systems, depth sensors (e.g., Kinect), wearable devices, or video-based pose estimation.

\textbf{Motion capture systems.} 
In the early stages of skeleton-based gait emotion recognition research, motion capture systems—particularly the VICON system—were widely used to acquire high-fidelity gait data. Omlor et al. \cite{OMLOR20071938} were among the first to utilize the VICON system to collect gait data from participants. They introduced a nonlinear source separation technique to extract spatiotemporal primitives from joint-angle trajectories of complex full-body movements, with a particular focus on patterns associated with emotional gait. Venture et al. \cite{5626404, venture2014recognizing} first conducted psychological experiments to investigate the factors influencing human perception of emotional gaits. They subsequently used feature vectors and Principal Component Analysis (PCA) to demonstrate the feasibility of numerical emotion recognition and proposed a similarity index–based algorithm for classification. Their experiments employed both 6-degree-of-freedom (DOF) and 12-DOF models.

Karg et al. \cite{5439949, karg2009comparison, karg2009two} extracted gait-related features such as walking speed, stride length, and the minimum, average, and maximum values of key joint angles (e.g., neck, shoulder, and chest). They applied dimensionality reduction techniques, including PCA, to manage feature complexity before employing machine learning models for emotion classification. In a related study, Daoudi et al. \cite{Daoudi2017} transformed joint position and velocity data into covariance matrices, which were then mapped onto the non-linear Riemannian manifold of Symmetric Positive Definite (SPD) matrices. Emotion classification was performed by computing geodesic distances and geometric means on the manifold.

\textbf{Depth sensors.} 
In 2010, the release of Microsoft’s depth-sensing camera introduced a novel technological approach for gait data acquisition and analysis. Researchers from the Chinese Academy of Sciences designed an experiment in which participants viewed emotion-inducing video clips, followed by walking sessions recorded using the Kinect v2 sensor \cite{li2016identifying, li2016emotion}. A Fourier transform was applied to extract 168 frequency-domain features to differentiate among happy, sad, and neutral emotional states \cite{li2016emotion}. Building on this work, Li et al. \cite{li2016identifying} further incorporated time-domain features such as stride length and gait cycle duration, resulting in a significant improvement in classification performance.

In a related study \cite{8482086}, geometric and kinematic features were extracted from human gait data. These included body-related features (e.g., head inclination angle, joint flexion angle), effort-related features (e.g., kinetic energy, average joint velocity), shape-related metrics (e.g., density index), and spatial descriptors (e.g., body contraction index, symmetry). A total of 17 features were extracted, processed through vector quantization and normalization. A binary chromosome–based genetic algorithm was then used to select optimal feature subsets for four expert models, thereby enhancing the emotion recognition performance of each model.

\textbf{Wearable devices.} 
With the widespread adoption of smartphones and wearable devices, the built-in sensors and cameras of these technologies have provided more accessible and convenient means for gait data collection. Zhang et al. \cite{10.7717/peerj.2258} and Cui et al. \cite{cui2016emotion} employed a custom-designed smart wristband to capture 3D acceleration data from various joints, including the right wrist and ankle. They computed time-domain features such as skewness, kurtosis, and standard deviation, and further extracted both time- and frequency-domain features using power spectral density and Fast Fourier transform (FFT) to classify three emotional states.

Similarly, Quiroz et al. \cite{info:doi/10.2196/10153, quiroz2017emotion} had participants wear smartwatches while walking along a 250-meter S-shaped corridor. Their study focused on inferring individuals' emotional states from sensor data collected via smartwatches, examining the relationship between motion sensor–captured gait patterns and emotional expressions. They extracted common statistical features from accelerometer data—including mean, standard deviation, maximum, and minimum values—and applied machine learning algorithms such as Random Forest and Logistic Regression for emotion classification.

\textbf{RGB Camera.} 
In addition to using smartphone sensors, Chiu et al.\cite{8480374} leveraged smartphone cameras to record participants’ gait videos and applied OpenPose\cite{cao2017realtime} to extract skeletal data. By analyzing Euclidean distance features, angular features, and velocity-based features derived from the skeletons, they successfully performed emotion classification based on gait.

\begin{table}[tp]
	\setlength{\tabcolsep}{1mm}
	\renewcommand\arraystretch{1.1}
	\centering
	\scriptsize  
	\caption{Emotion recognition from gait using deep learning methods}
	\label{Gait_DL_List}
	\begin{threeparttable}
		\begin{tabular}{@{}p{1.6cm} p{1.8cm} p{4.2cm} p{1.8cm} p{0.9cm} p{2.4cm}@{}}
			\toprule
			Study & Dataset & Feature & Classifier & Protocol & Accuracy \\ \midrule
			
			\rowcolor{gray!12}
			Venture et al. \cite{venture2014recognizing} 
			& 4 actors / 100 samples 
			& Lower torso movement (6DOF), waist rotations (3DOF), head inclinations (3DOF) 
			& Similarity Index 
			& LOSO 
			& 69\% \\
			
			Karg et al. \cite{5439949} 
			& 13 actors / 1300 samples 
			& Statistical params: velocity, stride length; PCA, KPCA, LDA, GDA; eigenposture via Fourier 
			& NB / NN / SVM 
			& LOSO 
			& 60\%-69\% \\
			
			\rowcolor{gray!12}
			Daoudi et al. \cite{Daoudi2017} 
			& BME 
			& Cov. matrix of posture/velocity vectors 
			& KNN 
			& LOSO 
			& 71.12\% \\
			
			Li et al. \cite{li2016emotion} 
			& 59 students 
			& 42 main frequencies + phases (Fourier) 
			& NB / RF / SVM / SMO 
			& 10-fold 
			& 51.69\%-80.51\% \\
			
			\rowcolor{gray!12}
			Li et al. \cite{li2016identifying} 
			& 59 students 
			& 44 time features, 2880 freq features via DFT of 6 joints 
			& LDA / NB / DT / SVM 
			& N/A 
			& 46\%-88\% \\
			
			Ferdous et al. \cite{8482086} 
			& 7 subjects 
			& 17 body/effort/shape/space features, quantized and normalized 
			& SVM / KNN / LDA / DT 
			& LOSO 
			& 62.86\%-74.29\% \\
			
			\rowcolor{gray!12}
			Zhang et al. \cite{10.7717/peerj.2258} 
			& 123 students 
			& Temporal (skew, kurtosis), freq (PSD), time-freq (FFT) features 
			& DT / SVM / RF 
			& 10-fold 
			& 56.60\%-81.20\% \\
			
			Chiu et al. \cite{8480374} 
			& 11 students 
			& L2, angular, speed features 
			& SVM / MLP / DT / NB / RF / LR 
			& 12-fold 
			& 53.3\%-62.1\% \\
			
			\bottomrule
		\end{tabular}
		\begin{tablenotes}
			\footnotesize
			\item SMO: Sequential Minimal Optimization; LDA: Linear Discriminant Analysis
		\end{tablenotes}
	\end{threeparttable}
\end{table}

\subsubsection{Feat2Net Approaches}

Since 2018, research on gait-based emotion recognition using skeletal data has entered a new phase characterized by the integration of handcrafted features with deep learning techniques.

Randhavane et al. \cite{randhavane2019identifying} employed LSTM networks to extract gait features, followed by emotion classification using traditional machine learning algorithms such as Support Vector Machines (SVM) and Random Forests (RF). Bhatia et al. \cite{bhatia2022motion} extracted handcrafted features, including joint angles and inter-joint distances, and used LSTM networks for emotion classification. Zhang et al.~\cite{ZHANG2024129600} proposed a hierarchical attention-based neural network (MAHANN) for gait emotion recognition. Their framework extracts motion features—such as the relative positions and velocities of 21 joints and walking speed—through a motion sentiment module, and action features—comprising five selected joint angles—via an action sentiment module. These features are then fused using a three-layer fully connected network for final emotion classification.

\subsubsection{FeatFusionNet Approaches}

Bhattacharya et al. \cite{bhattacharya2020step} further integrated handcrafted features with deep learning models by extracting 29 emotion-related features (e.g., stride length, joint angles) and concatenating them with the final layer of a Spatial-Temporal Graph Convolutional Network (ST-GCN) to enhance classification performance. In another study, they proposed a semi-supervised framework based on an autoencoder, which incorporates hierarchical attention pooling and latent embedding learning to perform emotion recognition \cite{bhattacharya2020take}.

Sun et al. \cite{SUN20221} extracted features such as joint angles and acceleration, fused visual information with raw skeleton data, and employed a bidirectional LSTM-based classifier to recognize four emotion categories. Hu et al. \cite{hu2022tntc} encoded skeleton joints and affective features into image representations, applied a two-stream CNN to extract features, and used a Transformer-based Complementarity Module (TCM) to capture long-range dependencies by leveraging complementary information between the two streams.

Zhang et al. \cite{zhu2024temporal, zhang2024ttgcn} utilized a GCN to model skeletal dynamics while concurrently employing a CNN to process handcrafted features; the outputs of both streams were fused at the decision level to enhance classification accuracy. Zhai et al. \cite{Zhai2024Looking} proposed a dual-stream framework combining posture flow and motion flow. The posture flow applied regression constraints based on handcrafted features to embed prior emotional knowledge into the deep model, while the motion flow constructed a high-order velocity–acceleration relational graph to capture emotional intensity. The final classification was achieved by fusing the outputs of both streams.

\subsubsection{End2EndNet Approaches}

With the advancement of deep learning techniques, there has been a gradual shift from traditional feature engineering toward end-to-end learning approaches in gait-based emotion recognition. Instead of relying on manually extracted temporal and frequency-domain features, recent studies focus on directly modeling the mapping between gait and emotional states using architectures such as Transformers and Graph Convolutional Networks (GCNs). Since 2020, end-to-end approaches have become the predominant paradigm for emotion recognition based on gait skeleton data.

\textbf{CNN-based models.}
Narayanan et al. \cite{narayanan2020proxemo} encoded multi-view skeleton data into pseudo-images by embedding 3D skeleton sequences into 244$\times$244 RGB images. Specifically, the $Z$, $Y$, and $X$ coordinates of skeletal joints at each time step were mapped to the $R$, $G$, and $B$ channels, respectively. A convolutional neural network (CNN), combined with temporal Transformers, was then used to extract cross-modal emotional features.

\textbf{GCN-based models.}
Zhuang et al. \cite{GCSN2021} proposed an extended joint connectivity scheme that incorporates a root-node fully connected strategy along with a contraction-denoising module, resulting in a 12.6\% improvement in model performance. Lu et al. \cite{lu2023eipc} conducted an in-depth investigation into the impact of joint topology on emotion recognition and introduced an optimized joint connectivity design to enhance classification accuracy. Sheng et al. \cite{sheng2021multi} developed an Attention-enhanced Spatiotemporal Graph Convolutional Network (AE-STGCN) with an encoder-decoder architecture, enabling simultaneous modeling of spatial dependencies and temporal dynamics. Their framework supports multi-task learning for joint identity recognition and emotion classification. Yin et al. \cite{YIN2024110117} introduced the Multi-Scale Adaptive Graph Convolution Network (MSA-GCN) for gait emotion recognition. Their model integrates Adaptive Selective Spatio-Temporal Graph Convolution (ASST-GCN) to dynamically select convolution kernels based on emotional context and applies cross-scale mapping interaction to fuse multiscale information. Chen et al. \cite{Chen2023STAGCN} presented the Spatial-Temporal Adaptive Graph Convolutional Network (STA-GCN), which addresses the limitations of conventional models in capturing implicit joint relationships and rigid multi-scale temporal feature aggregation. This is achieved through dedicated spatial and temporal feature learning modules.

\textbf{Transformer-based models.}
Zeng et al. \cite{zeng2025gaitcycformer} introduced GaitCycFormer, a Transformer-based framework that incorporates cycle position encoding along with a bi-level architecture consisting of Intra-cycle and Inter-cycle Transformers. This design enables the model to effectively capture both local intra-cycle and global inter-cycle temporal features for gait-based emotion recognition.

\subsubsection{Unsupervised Approaches}

Due to the limited availability of gait emotion datasets, some researchers have begun exploring unsupervised emotion recognition methods. The primary strategy involves training an encoder to extract emotion-relevant gait features without labeled supervision, followed by evaluating the quality of the learned representations through a series of downstream tasks.

Lu et al. \cite{Lu_2023} proposed a self-supervised contrastive learning framework for gait-based emotion recognition, aiming to address the challenges of limited gait diversity and semantic consistency in existing methods. The framework incorporates two core components: (1) Ambiguity Contrastive Learning, which generates ambiguous samples by modifying gait speed and joint angles, and integrates them into the memory bank to enrich semantic diversity; (2) Cross-coordinate Contrastive Learning ($C^{3}L$), which performs contrastive learning between Cartesian and Spherical coordinate systems to leverage complementary representations for improved semantic invariance.

Similarly, Song et al. \cite{song2024self} presented a self-supervised contrastive framework that introduces selective strong augmentation, including techniques such as upper-body jitter and random spatiotemporal masking, to generate diverse positive samples and promote robust feature learning. They further designed a Complementary Feature Fusion Network (CFFN) to integrate topological features from the graph domain (via ST-GCN) with global adaptive features from the image domain (via an adaptive frequency filter), thereby enhancing representational capacity. To ensure distributional consistency between general and strongly augmented samples, a distributional divergence minimization loss is applied.

\subsubsection{Evaluation of Methods on Public Datasets}
To provide a clear and systematic comparison of method performance, we summarize the results of representative models evaluated on two widely used gait-based emotion recognition datasets: E-Gait and EMOGAIT. The comparative results are presented in Tables~\ref{Dataset_EGait_List} and~\ref{Dataset_EMOGAIT_List}.

On the E-Gait dataset, GCN-based approaches such as MSA-GCN \cite{YIN2024110117} and BPM-GCN \cite{Zhai2024Looking} demonstrated strong performance, with MSA-GCN achieving the highest reported accuracy of 93.51\%. CNN- and Transformer-based models, including MAHANN \cite{ZHANG2024129600} and GaitCycFormer \cite{zeng2025gaitcycformer}, also performed competitively, indicating the effectiveness of deep learning models in capturing both local and global gait dynamics.

On the EMOGAIT dataset, recent GCN variants—such as T2A \cite{zhu2024temporal} and TT-GCN \cite{zhang2024ttgcn}—achieved over 90\% accuracy, further confirming the efficacy of graph-based temporal modeling for emotion recognition from gait.

\begin{table}[tp]  
	\setlength{\tabcolsep}{1mm}
	\renewcommand\arraystretch{1.2}
	\centering
	\fontsize{9}{12}\selectfont
	\caption{Method comparison on E-Gait dataset}  
	\label{Dataset_EGait_List}
	\begin{threeparttable}  
		\begin{tabular}{@{}llll@{}}
			\toprule
			Study                                                            & Backbone          & Protocol  & Accuracy    \\ \midrule
			
			\rowcolor{gray!12}
			TAEW \cite{bhattacharya2020take} & RNN               & Hold-out  & 84.00\%           \\
			STEP \cite{bhattacharya2020step} & GCN               & Hold-out  & 82.15\%        \\
			
			\rowcolor{gray!12}
			MSA-GCN \cite{YIN2024110117}                 & GCN               & Hold-out  & 93.51\%        \\ 
			T2A \cite{zhu2024temporal}             & GCN               & 5-fold    & 82.91\% \\
			
			\rowcolor{gray!12}
			TT-GCN \cite{ zhang2024ttgcn}             & GCN               & 5-fold    & 80.11\%  \\
			BPM-GCN \cite{Zhai2024Looking}              & GCN               & 5-fold    & 88.94\%  \\
			
			\rowcolor{gray!12}
			EIPC \cite{lu2023eipc}                     & GCN               & 5-fold    & 82.25\%        \\
			G-GCSN \cite{GCSN2021}                   & GCN               & 10 - fold & 81.50\%        \\
			
			\rowcolor{gray!12}
			STA-GCN \cite{Chen2023STAGCN}               & GCN               & N/A       & 85.80\%        \\
			MAHANN \cite{ZHANG2024129600}             & CNN               & Hold-out  & 93.40\%        \\
			
			\rowcolor{gray!12}
			VFL \cite{SUN20221}                      & CNN               & Hold-out  & 89.29\%        \\
			Proxemo \cite{narayanan2020proxemo}    & CNN               & 5-fold    & 80.01\%  \\
			
			\rowcolor{gray!12}
			TNTC \cite{hu2022tntc}                            & Transfromer       & 5-fold    & 85.97\%  \\
			Gaitcycformer \cite{zeng2025gaitcycformer}        & Graph-Transformer & Hold-out  & 86.30\%        \\ \bottomrule
		\end{tabular}
	\end{threeparttable} 
\end{table}

\begin{table}[tp]
	\centering
	\fontsize{9}{12}\selectfont
	\setlength{\tabcolsep}{2mm}
	\renewcommand\arraystretch{1.2}
	\caption{Method comparison on EMOGAIT dataset}
	\label{Dataset_EMOGAIT_List}
	
	
	\begin{threeparttable}
		\begin{tabular}{llll}
			\toprule
			Study & Backbone & Protocol & Accuracy \\
			\midrule
			\rowcolor{gray!12}
			T2A \cite{zhu2024temporal}   & GCN & 5-fold   & 91.87\% \\
			TT-GCN \cite{zhang2024ttgcn} & GCN & 5-fold   & 90.25\% \\
			
			\rowcolor{gray!12}
			AT-GCN \cite{sheng2021multi} & GCN & Hold-out & 86.80\% \\
			\bottomrule
		\end{tabular}
	\end{threeparttable}
\end{table}

\subsection{Methods Comparison and Analysis}\label{sec4_3}
Despite differences in data sources and classification algorithms, most traditional methods share common strategies for extracting emotion-related features from skeletal data. Through a detailed investigation of numerous representative studies, we observed that frequently used features include joint velocities, inter-joint angles, and relative distances—parameters that effectively reflect subtle emotional variations in posture and movement. We summarized these commonly adopted features in Table \ref{Dataset_Features_List}.

\begin{table*}[tp]
	\centering
	\fontsize{10}{12}\selectfont
	\setlength{\tabcolsep}{1mm}
	\renewcommand\arraystretch{1.2}
	\caption{Common posture-emotion feature list}
	\label{Dataset_Features_List}
	
	\begin{adjustbox}{max width=\textwidth}
		\begin{threeparttable}
			\begin{tabular}{L{3.2cm} C{3.2cm} L{12.6cm}}
				\toprule
				\multicolumn{2}{c}{Feature Type} & Detailed Description \\
				\midrule
				\multirow{8}{*}{Postural Features} &
				\multirow{3}{*}{Angle Features} &
				\cellcolor{gray!15}Left shoulder–neck–right shoulder, left shoulder–neck–left elbow, \\[-0.2em]
				& & Left hip–waist–left knee$^{\ast}$, left shoulder–left elbow–left hand$^{\ast}$ \\[-0.2em]
				& & \cellcolor{gray!15}left hip–left knee–left foot$^{\ast}$, Head–neck–torso \\ \cmidrule(l){2-3}
				& \multirow{2}{*}{Distance Features} &
				Left hand–torso$^{\ast}$, left hand–left shoulder$^{\ast}$, left hand–left hip$^{\ast}$ \\[-0.2em]
				& & \cellcolor{gray!15}Left hand–neck$^{\ast}$, left elbow–torso$^{\ast}$, left foot–right foot \\ \cmidrule(l){2-3}
				& \multirow{3}{*}{Area Features} &
				Left hand–neck–right hand, left shoulder–neck–right shoulder \\[-0.2em]
				& & \cellcolor{gray!15}Left hand–hip–right hand, left elbow–neck–right elbow \\[-0.2em]
				& & Left foot–waist–right foot, left knee–neck–right knee \\ \cmidrule(l){1-3}
				\multirow{2}{*}{Motion Features} &
				Velocity Features &
				\cellcolor{gray!15}Head, hands, shoulders, knees, feet \\ \cmidrule(l){2-3}
				& Acceleration Features &
				Head, hands, shoulders, knees, feet \\
				\bottomrule
			\end{tabular}
			
			\begin{tablenotes}
				\footnotesize
				\item[$\ast$] The corresponding feature is also computed on the right side (e.g.\ for left shoulder–left elbow–left hand, compute right shoulder–right elbow–right hand).
			\end{tablenotes}
		\end{threeparttable}
	\end{adjustbox}
\end{table*}

While deep learning methods are widely regarded as end-to-end solutions, our analysis reveals that many still rely on handcrafted features—particularly in cases where data scarcity or interpretability is a concern. For example, several Feat2Net and FeatFusionNet models incorporate domain-specific features such as motion energy or joint angle dynamics as inputs to neural networks. This hybrid design reflects a transitional stage in the evolution of deep learning approaches toward fully automated representation learning.

Overall, graph-based approaches are playing an increasingly dominant role among deep learning methods, particularly those augmented with temporal attention mechanisms or multi-scale modeling strategies. CNN- and Transformer-based models also demonstrate significant potential, especially when designed to capture fine-grained motion patterns or long-range dependencies. These trends reflect a broader shift toward end-to-end, data-driven frameworks capable of learning expressive and discriminative representations directly from skeletal gait sequences.

Meanwhile, although the method proposed by Lu et al. \cite{Lu_Chen_Liang_Tan_Zeng_Hu_2025} trails slightly in classification accuracy, it introduces a unique advantage: the ability to recognize emotions while simultaneously generating descriptive textual explanations. This LLM-based approach opens a promising direction for future research, enabling emotion recognition systems not only to detect emotional states but also to offer human-interpretable insights that enhance transparency and user trust.

\section{Task-Specific Applications}\label{sec_5}

\subsection{Depression Detection Using Skeleton Data}\label{sec_5_1}

Emotion is typically considered a short-term mental state, whereas depression is a long-term and multifaceted psychological condition. Despite these distinctions, the two are closely interconnected \cite{deligianni2019emotions}. Individuals experiencing depression often suffer from persistent negative emotions and extended periods of emotional decline. As a result, researchers have sought to extend the insights gained from gait-based emotion recognition to the domain of gait-based depression detection, aiming to explore the potential of gait features as indicators of depressive symptoms.

A research team from the Institute of Psychology, Chinese Academy of Sciences \cite{zhao2019see, yuan2019depression, wang2021detecting} analyzed gait characteristics in the frequency domain using methods such as Fast Fourier Transform (FFT) and Hilbert–Huang Transform (HHT), establishing a mapping between gait features and levels of depression. Lu et al. \cite{Lu2021A, Lu2022MM} investigated gait patterns in individuals with depression and proposed joint energy features that effectively differentiated between depressed patients and healthy controls. Fang et al. \cite{fang2019depression}, from the Shenzhen Institutes of Advanced Technology, Chinese Academy of Sciences, extracted 12 time-domain gait features—such as walking speed, stride width, stride length, and vertical head motion—and analyzed their correlations with depressive symptoms.

Building on these earlier works, Wang et al. \cite{wang2020gait} integrated time-domain, frequency-domain, and spatial-geometric features to develop a novel gait-based depression assessment algorithm. In addition, Yang et al. \cite{Yang2022} applied various data augmentation strategies to skeleton sequences to examine how different forms of skeleton data influence depression recognition performance. Shao et al. \cite{shao2021multi} further enhanced model robustness by incorporating video data alongside skeleton sequences and introducing a fusion mechanism between skeletal and gait contour features.


Based on skeletons extracted from video dataset \cite{liu2024depression}, Li et al. \cite{11076140} proposes a novel Spatio-Temporal Multi-granularity Network (STM-Net) for skeleton-based depression risk recognition, incorporating a Multi-Grain Temporal Focus (MTF) module and a Multi-Grain Spatial Focus (MSF) module to capture dynamic temporal information and spatial features in gait patterns. 

\subsection{Autism Detection Using Skeleton Data}\label{sec_5_2}

Autism spectrum disorder (ASD) is a long-term neurodevelopmental condition characterized by difficulties in social communication and the presence of restricted, repetitive behaviors. Emerging studies have suggested that individuals with ASD often exhibit amplified emotional responses and impaired emotional regulation, which may be reflected in distinct physical behaviors compared to neurotypical individuals \cite{mazefsky2013role}. As a result, researchers have begun to investigate the identification of ASD through behavioral cues—particularly gait patterns—as a non-invasive means of understanding and detecting autism-related traits. This line of research holds promise for contributing to early diagnosis and intervention strategies.

Study \cite{al2020generating} was the first to introduce a framework for capturing and constructing a 3D gait and full-body motion dataset of children with ASD using the Kinect v2 depth camera, providing valuable resources for behavioral analysis and ASD-related research. Zhang et al. \cite{zhang2021application} employed the OpenPose algorithm to extract initial skeleton data from video recordings. A skeleton distance–matching algorithm was then used for multi-person tracking to associate the skeleton data of multiple ASD children within the same scene. Finally, an LSTM network was applied to classify the denoised skeleton sequences and automatically extract temporal features.

Zahan et al.\cite{zahan2023human} utilized a GCN with angle embedding to capture spatiotemporal features from skeletal data and incorporated skeleton image representations (Skepxels) alongside a Vision Transformer (ViT) for auxiliary training. Their findings indicate that children with autism exhibit atypical gait characteristics, such as greater joint angle variability and increased gait asymmetry. Yang et al. \cite{yang2025body} used HRNet \cite{sun2019deep} to extract skeletal keypoint coordinates and applied the PoTion algorithm to generate joint motion trajectory maps. Their study revealed that children with ASD tend to exhibit broader movement ranges and more irregular motion patterns compared to neurotypical peers.

\subsection{Abnormal Behavior Detection Using Skeleton Data}\label{sec_5_3}

Abnormal behavior detection and emotion recognition are two interrelated yet distinct tasks. Emotion recognition aims to identify short-term affective states, whereas abnormal behavior detection focuses on recognizing unusual or unexpected patterns that may signal underlying psychological or behavioral conditions \cite{10453042}. Despite this distinction, the two domains are closely connected, as many abnormal behaviors are either driven by or accompanied by emotional changes—for example, heightened anxiety may manifest as restlessness or avoidance behavior.

\textbf{RNN-based models.}
In the early stages of research, recurrent neural networks (RNNs) were the predominant technique for addressing these tasks. In~\cite{zhou2020skeleton}, a monocular fixed-camera depth estimation algorithm was used to convert 2D skeletons into 3D representations. A self-attention–based spatiotemporal convolutional neural network (ST-CNN) was employed to extract local spatiotemporal features, while an attention-enhanced LSTM (ATT-LSTM) focused on key frames to capture global temporal dynamics. Morais et al.~\cite{morais2019learning} decomposed human skeletal movements into global body motion and local body posture components, which were modeled using a novel Message-Passing Encoder-Decoder Recurrent Network (MPED-RNN). This model consists of two interacting RNN branches—one for global and one for local features—that exchange information through cross-branch message passing at each time step. The network is trained using both reconstruction and prediction losses.

\textbf{GCN-based models.}
Subsequently, Graph Convolutional Networks (GCNs) gradually emerged as the dominant paradigm for skeleton-based abnormal behavior detection. Markovitz et al.~\cite{markovitz2020graph} introduced the Graph Embedded Pose Clustering (GEPC) method for anomaly detection, representing human poses as spatio-temporal graphs. Their framework employs a Spatio-Temporal Graph Convolutional Autoencoder (STGCAE) to embed pose graphs, applies deep embedded clustering to produce soft-assignment vectors, and utilizes a Dirichlet Process Mixture Model (DPMM) to compute normality scores.

Liu et al.~\cite{liu2021self} proposed a Spatial Self-Attention Augmented Graph Convolution (SAA-Graph) module that integrates improved spatiotemporal graph convolutions with Transformer-style self-attention to capture both local and global joint information. Their architecture uses a SAA-STGCAE for feature extraction, followed by deep embedded clustering and DPMM for anomaly scoring.

Flaborea et al.~\cite{flaborea2024contracting} introduced COSKAD, a method that encodes skeletal motion using a Space-Time-Separable Graph Convolutional Network (STS-GCN). The model projects skeletal embeddings into multiple latent spaces—Euclidean, spherical, and hyperbolic—and detects anomalies by minimizing distances from a learned central point in each space.

Karami et al.~\cite{karami2025graph} presented GiCiSAD, a Graph-Jigsaw Conditioned Diffusion Model for skeleton-based video anomaly detection. The framework consists of three novel modules: (1) a Graph Attention–based Forecasting module to model spatio-temporal dependencies, (2) a Graph-level Jigsaw Puzzle Maker to highlight subtle region-level discrepancies, and (3) a Graph-based Conditional Diffusion model for generating diverse human motion patterns to aid anomaly detection.

\textbf{Pretrained and prompt-guided models.}
More recently, an increasing number of studies have begun to incorporate pretrained weights and large-scale models to enhance recognition performance in skeleton-based abnormal behavior detection. Sato et al.~\cite{sato2023prompt} proposed a prompt-guided zero-shot framework for anomaly action recognition, addressing several limitations of traditional skeleton-based methods—namely, the reliance on target-domain training, sensitivity to skeleton errors, and the scarcity of normal sample annotations. Their approach employs a PointNet-based permutation-invariant extractor to enable sparse feature propagation and improve robustness against skeletal noise. The framework also integrates cosine similarity between skeleton features and text embeddings (derived from user-provided abnormal action prompts) into anomaly scores via contrastive learning, thereby indirectly incorporating knowledge of normal behavior.

Liu et al.~\cite{liu2024language} introduced the Contrastive Language-Skeleton Pre-training framework (SkeletonCLSP), which leverages large language models to enhance skeleton-based action recognition through three key mechanisms: semantic compensation, cross-modal feature integration, and anomaly correction. This framework bridges the semantic gap between textual and skeletal modalities, enabling more robust and generalizable recognition of abnormal actions.

\section{Challenges and Future Research Directions}\label{sec_6}




\subsection{Constructing Diversified Datasets} \label{sec_6_1}

Despite growing interest in recognizing emotional body expressions from 3D skeleton data, the availability of high-quality, large-scale datasets remains limited. Most existing datasets are relatively small in scale and are predominantly annotated using discrete emotion categories (e.g., happiness, anger, sadness), as summarized in Table~\ref{Dateset_List}. Based on this limitation, several key aspects need to be addressed in dataset development:

\begin{itemize}
    \item \textbf{Enriching Annotation Schemes}: Most existing datasets are annotated with discrete emotion categories (e.g., happiness, anger, sadness), but these may not capture the full complexity of emotional body movements. There is a need to incorporate dimensional emotion models, such as valence–arousal or PAD, for more nuanced emotional representations, especially in applications like abnormal behavior detection and context-aware interventions \cite{dahmane2020multimodal}. Moreover, dataset development should not only focus on scale and diversity but also adopt richer, multi-modal annotation schemes that integrate both categorical and dimensional labels. Including variables such as age, gender, and cultural background could enhance the generalizability and fairness of recognition systems across diverse populations \cite{kleinsmith2006cross}.

    \item \textbf{Expanding Data Collection Scenarios}: Most existing public datasets for emotional body expressions are collected in highly controlled laboratory environments, as shown in Table~\ref{Dateset_com}. While such settings ensure clean data and consistent annotations, they often lack the robustness in real-world applications. It is crucial to extend data collection efforts from highly controlled laboratory environments to more naturalistic, real-world settings. This includes acquiring 3D skeleton data in diverse environments such as workplaces, public spaces, classrooms, or healthcare settings, where emotional expressions emerge in response to complex social and environmental stimuli \cite{guo2024development}.
    
    \item \textbf{Exploring Generative Techniques}: Future research should explore generative techniques for data augmentation, such as semi-supervised, unsupervised \cite{zhang2025skeleton}, or synthetic data generation \cite{Yang2022}. These approaches can help create richer, more diverse datasets and reduce reliance on large manually labeled datasets while improving model generalizability and scalability.
\end{itemize}

\subsection{Improving Model Performance}\label{sec_6_3}

Based on the analysis of existing skeleton-based emotion recognition methods, there remains significant room for improvement in several aspects of model design.

\begin{itemize}
	\item \textbf{Accuracy}: Achieving high recognition accuracy remains a core goal. Emotion expressions in body movements can be subtle, ambiguous, or vary across individuals and contexts, making it difficult for models to capture discriminative features reliably. Complex emotions or overlapping expressions further compound the difficulty.
	\item \textbf{Generalization}: Models trained on a specific dataset or population often struggle to generalize to unseen users, diverse cultural backgrounds, or different recording environments. Domain shift—caused by variations in pose quality, motion style, or camera viewpoints—can significantly degrade performance in real-world applications.
	\item \textbf{Interpretability}: Black-box models offer limited insight into why a particular emotion was predicted. Enhancing model interpretability is important for debugging, trust, and downstream applications such as emotional reasoning or ethical AI. Approaches such as attention mechanisms, saliency mapping, or symbolic reasoning may help uncover the decision logic behind model outputs.
\end{itemize}

\subsection{Building End-to-End and Efficient Emotion Recognition Frameworks}\label{sec_6_4}

Real-time emotion recognition is critical for applications such as human–robot interaction, public safety surveillance, and affective computing on edge devices. However, current skeleton-based emotion recognition pipelines often rely on multi-stage processes—either acquiring skeletal data from dedicated sensors or extracting skeletons from RGB videos using pose estimation algorithms~\cite{shen2019emotion, mingming2020emotion}. These intermediate steps not only increase system complexity and latency but also introduce noise and error propagation, ultimately degrading recognition accuracy and computational efficiency.

To overcome these limitations, future research should explore the development of end-to-end frameworks that map raw input data (e.g., video or depth frames) directly to emotional states, without a heavy dependence on intermediate skeletal representations or extensive preprocessing. Such approaches could significantly reduce computational overhead and enable more streamlined deployment, particularly in resource-constrained environments~\cite{paiva2025skelett, luo2020arbee}.

In addition, designing lightweight models with strong generalization capabilities and low memory or computational requirements is essential for real-time inference. This necessitates innovations in efficient model architectures, including transformer pruning, knowledge distillation, and edge-optimized neural operators, to balance performance and deployment feasibility.

\subsection{Expanding to Multi-person Emotion Recognition}\label{sec_6_5}

Most existing methods focus on single-person emotion recognition; however, many real-world applications—such as public safety, education, and social robotics—demand the ability to understand group-level emotions~\cite{li2025decoding}.

Recognizing collective affect requires modeling not only individual emotional states but also interpersonal dynamics and group context. This introduces a range of challenges, including occlusions in crowded scenes, variations in individual expressiveness, and the lack of dedicated multi-person skeleton-based emotion datasets.

Future research should investigate relational modeling approaches, such as graph or hypergraph representations, to capture both individual cues and group-level interactions. Such methods could enable the development of more socially aware, context-sensitive, and robust emotion recognition systems suitable for real-world, multi-agent environments.

\subsection{Expanding to Multimodal Emotion Recognition}\label{sec_6_6}
While 3D skeleton data captures rich information about body movements, emotional expressions are inherently multimodal, encompassing facial expressions, vocal cues, physiological signals, and even brain activity. Relying solely on skeletal data may constrain recognition accuracy, particularly for subtle or ambiguous emotional states.

Integrating additional modalities—such as audio, video, or EEG—can provide complementary cues and enhance system robustness~\cite{zhang2024deep, li2022eeg}. For instance, combining body gestures with speech prosody or facial expressions can substantially improve emotion recognition in naturalistic settings~\cite{yan2024multimodal}. EEG signals, which reflect internal affective processes, are especially valuable in scenarios where external emotional expressions are weak or intentionally suppressed~\cite{kim2024enhancing}.

However, multimodal fusion introduces several key challenges:
\begin{itemize}
	\item \textbf{Modality alignment and synchronization}, particularly when input signals have different temporal or spatial resolutions.
	\item \textbf{Increased model complexity and computational demands}, which may hinder real-time deployment.
	\item \textbf{Limited availability of large-scale, well-annotated multimodal emotion datasets}, impeding training and evaluation.
\end{itemize}


\subsection{Leveraging Large Models for Skeleton-Based Emotion Recognition}\label{sec_6_7}

Recent advances in large-scale models—such as ChatGPT~\cite{ChatGPT}, LLaVa~\cite{liu2024visual}, and Gemini~\cite{gemini15pro2024}—have opened new avenues for skeleton-based emotion recognition. However, directly applying these models to 3D skeleton sequences remains challenging due to the modality’s sparse, temporal, and structured nature. Unlike text or images, skeleton data requires careful preprocessing—such as transformation into token-like representations or pseudo-images—to align with the input formats expected by large-scale models~\cite{Lu_Chen_Liang_Tan_Zeng_Hu_2025}.

To better harness the capabilities of these models, future research could explore unified architectures that support not only emotion recognition but also understanding and generation. In particular, such models could be designed to:

\begin{itemize}
	\item \textbf{Interpret:} the underlying causes of emotional expressions by incorporating contextual and situational information.
	\item \textbf{Generate:} emotionally expressive body movements conditioned on target emotion labels or predefined scenarios.
	\item \textbf{Reason:} about the temporal evolution of emotional states in interactive or multi-agent environments.
\end{itemize}


\section{Conclusion}
This survey presents a comprehensive review of recent advances in emotion recognition based on 3D skeletal data, encompassing both posture-based and gait-based approaches. By examining data acquisition methods, publicly available datasets, and a range of technical strategies—from traditional handcrafted feature extraction to deep learning architectures and large-scale pretrained models—we offer a unified perspective on the evolution of this rapidly developing field.

Compared to facial expression or voice-based methods, skeleton-based emotion recognition offers distinct advantages, including robustness to environmental variations and improved privacy protection. These characteristics make it particularly well-suited for real-world applications such as healthcare monitoring, human–computer interaction, and public safety.

Despite considerable progress, several key challenges remain. These include the need for more diverse and ecologically valid datasets, improved generalization across users and environments, and enhanced interpretability of deep learning models. In addition, integrating multimodal signals (e.g., voice, facial expression, and physiological data) and leveraging large-scale pretrained models present promising avenues for future research.

We anticipate that the next wave of research will focus on developing unified, end-to-end frameworks that are lightweight, explainable, and capable of adapting to dynamic, multimodal environments. Such systems will be critical for enabling robust, scalable, and human-centered affective computing in real-world scenarios.

\bibliographystyle{IEEEtran}
\bibliography{sample-base}


\end{document}